\documentclass[twoside]{article}
\usepackage[accepted]{aistats2020}

\usepackage[T1]{fontenc}
\usepackage{aecompl}

\usepackage[round]{natbib}
\usepackage{xcolor}
\definecolor{mydarkred}{rgb}{0.6,0,0}
\definecolor{mydarkgreen}{rgb}{0,0.6,0}
\usepackage[colorlinks,
linkcolor=mydarkred,
citecolor=mydarkgreen]{hyperref}
\usepackage{url}

\usepackage{wrapfig}
\usepackage{graphicx}
\graphicspath{{figure/}}
\usepackage[skip=1ex]{caption}
\usepackage[skip=0ex]{subcaption}

\usepackage{amsmath,amsthm,amssymb,bm}
\usepackage{mathrsfs}
\usepackage{nccbbb}
\newtheorem{theorem}{Theorem}
\newtheorem{definition}[theorem]{Definition}
\newtheorem{lemma}[theorem]{Lemma}
\newtheorem{corollary}[theorem]{Corollary}

\usepackage{algorithm,algorithmic}
\usepackage{enumitem}
\setlist{nosep}
\usepackage{booktabs}
\usepackage{multirow}
\usepackage{multicol}

\DeclareMathOperator{\sign}{\mathrm{sign}}
\DeclareMathOperator*{\argmin}{\mathrm{arg\,min}}

\newcommand{\dif}{\mathrm{d}}
\newcommand{\pr}{\mathrm{Pr}}

\newcommand{\bE}{\mathbb{E}}
\newcommand{\bR}{\mathbb{R}}
\newcommand{\cA}{\mathcal{A}}
\newcommand{\cG}{\mathcal{G}}
\newcommand{\cF}{\mathcal{F}}
\newcommand{\cO}{\mathcal{O}}

\newcommand{\cX}{\mathcal{X}}
\newcommand{\fD}{\mathfrak{D}}
\newcommand{\fR}{\mathfrak{R}}

\newcommand{\ptr}{p_\mathrm{tr}}
\newcommand{\pip}{\pi_\mathrm{p}}

\newcommand{\pin}{\pi_\mathrm{n}}

\newcommand{\Xtr}{\cX_\mathrm{tr}}

\newcommand{\bXtr}{\overline{\cX}_\mathrm{tr}}
\newcommand{\Xp}{\cX_\mathrm{p}}
\newcommand{\Xn}{\cX_\mathrm{n}}

\newcommand{\hRpn}{\widehat{R}_\mathrm{pn}}

\newcommand{\hRuu}{\widehat{R}_\mathrm{uu}}

\newcommand{\hRui}{\widehat{R}_\mathrm{u}}
\newcommand{\hRuii}{\widehat{R}_\mathrm{u'}}
\newcommand{\hRnnuu}{\widehat{R}_\mathrm{cc}}

\newcommand{\hgcc}{\widehat{g}_\mathrm{cc}}

\newcommand{\Rp}{R_\mathrm{p}}
\newcommand{\Rn}{R_\mathrm{n}}

\newcommand{\hRp}{\widehat{R}_\mathrm{p}}
\newcommand{\hRn}{\widehat{R}_\mathrm{n}}

\newcommand{\Np}{{n_\mathrm{p}}}
\newcommand{\Nn}{{n_\mathrm{n}}}

\newcommand{\ellsig}{\ell_\mathrm{sig}}
\newcommand{\elllog}{\ell_\mathrm{log}}

\newcommand{\tabincell}[2]{\begin{tabular}{@{}#1@{}}#2\end{tabular}}

\begin{document}

\twocolumn[

\aistatstitle{Mitigating Overfitting in Supervised Classification from Two\\ Unlabeled Datasets: A Consistent Risk Correction Approach}

 \aistatsauthor{ Nan Lu$^{1,2}$ \quad Tianyi Zhang$^{1}$ \quad  Gang Niu$^{2}$ \quad  Masashi Sugiyama$^{2,1}$ }

 \aistatsaddress{ $^{1}$The University of Tokyo, Japan \qquad $^{2}$RIKEN, Japan} 

]


\begin{abstract}
The recently proposed unlabeled-unlabeled (UU) classification method allows us to train a binary classifier only from two \emph{unlabeled} datasets with different class priors. Since this method is based on the empirical risk minimization, it works as if it is a \emph{supervised} classification method, compatible with any model and optimizer. However, this method sometimes suffers from severe overfitting, which we would like to prevent in this paper. Our empirical finding in applying the original UU method is that overfitting often co-occurs with the empirical risk \emph{going negative}, which is not legitimate. Therefore, we propose to \emph{wrap} the terms that cause a negative empirical risk by certain \emph{correction functions}. Then, we prove the consistency of the corrected risk estimator and derive an estimation error bound for the corrected risk minimizer. Experiments show that our proposal can successfully mitigate overfitting of the UU method and significantly improve the classification accuracy.
\end{abstract}
\begin{figure*}[t]
    \centering
    \begin{minipage}[c]{0.19\textwidth}~\end{minipage}%
    \begin{minipage}[c]{0.2\textwidth}\centering\small Linear, $\ellsig$ \end{minipage}%
    \begin{minipage}[c]{0.2\textwidth}\centering\small Linear, $\elllog$ \end{minipage}%
    \begin{minipage}[c]{0.2\textwidth}\centering\small MLP, $\ellsig$ \end{minipage}%
    \begin{minipage}[c]{0.2\textwidth}\centering\small MLP, $\elllog$ \end{minipage}\\
    \begin{minipage}[c]{0.17\textwidth}\centering\small MNIST dataset\\SGD optimizer \end{minipage}%
    \begin{minipage}[c]{0.8\textwidth}
        \includegraphics[width=0.25\textwidth]{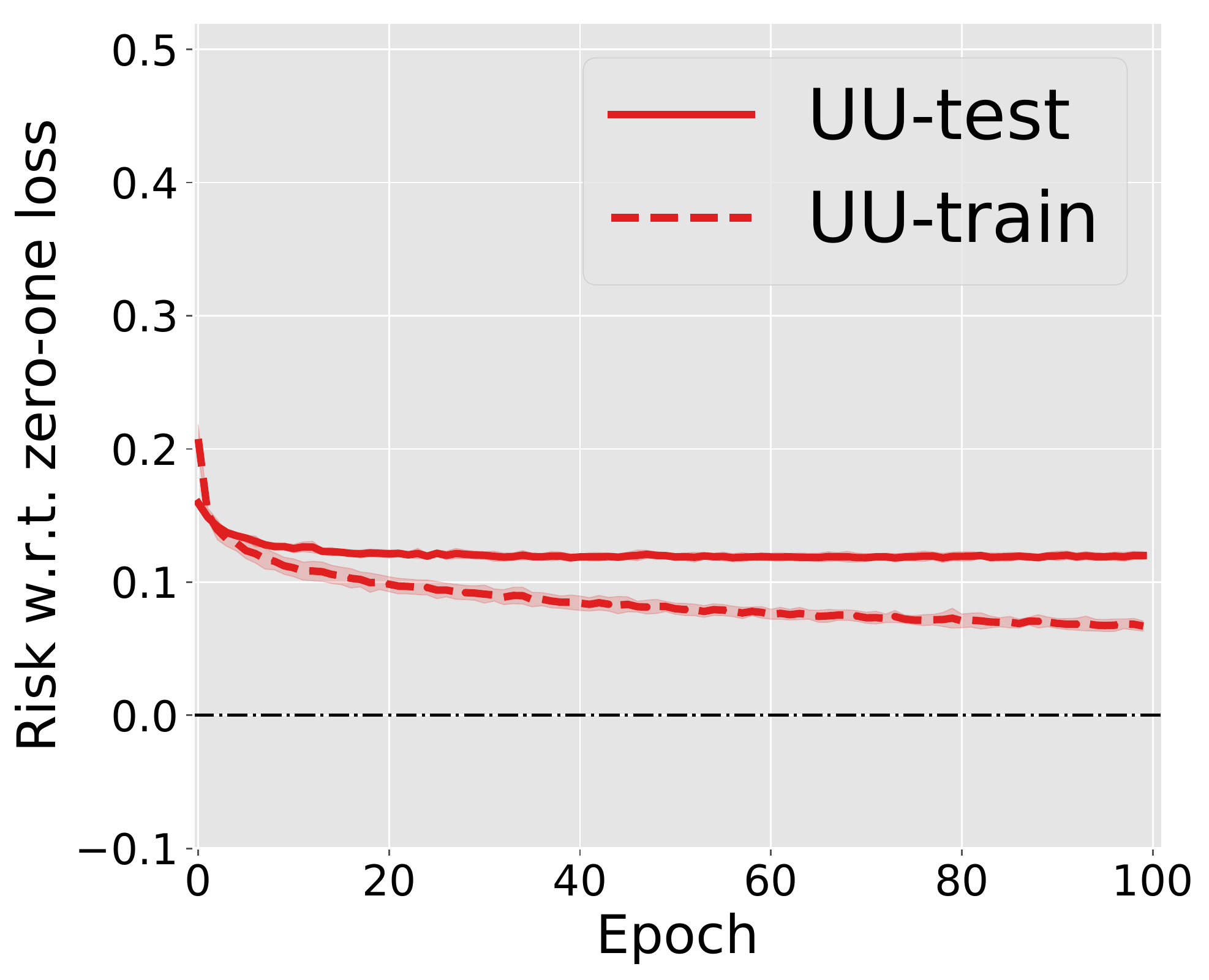}%
        \includegraphics[width=0.25\textwidth]{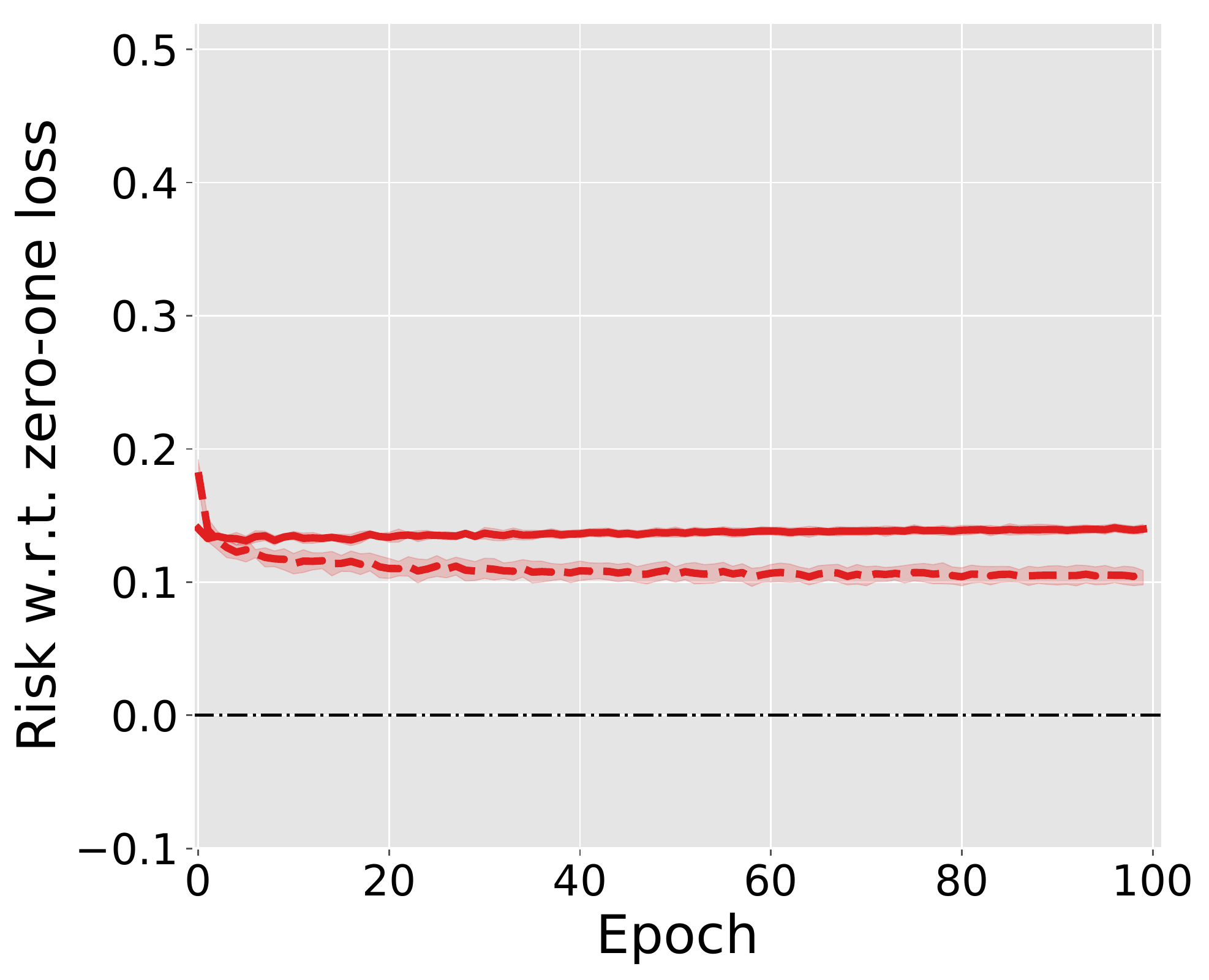}%
        \includegraphics[width=0.25\textwidth]{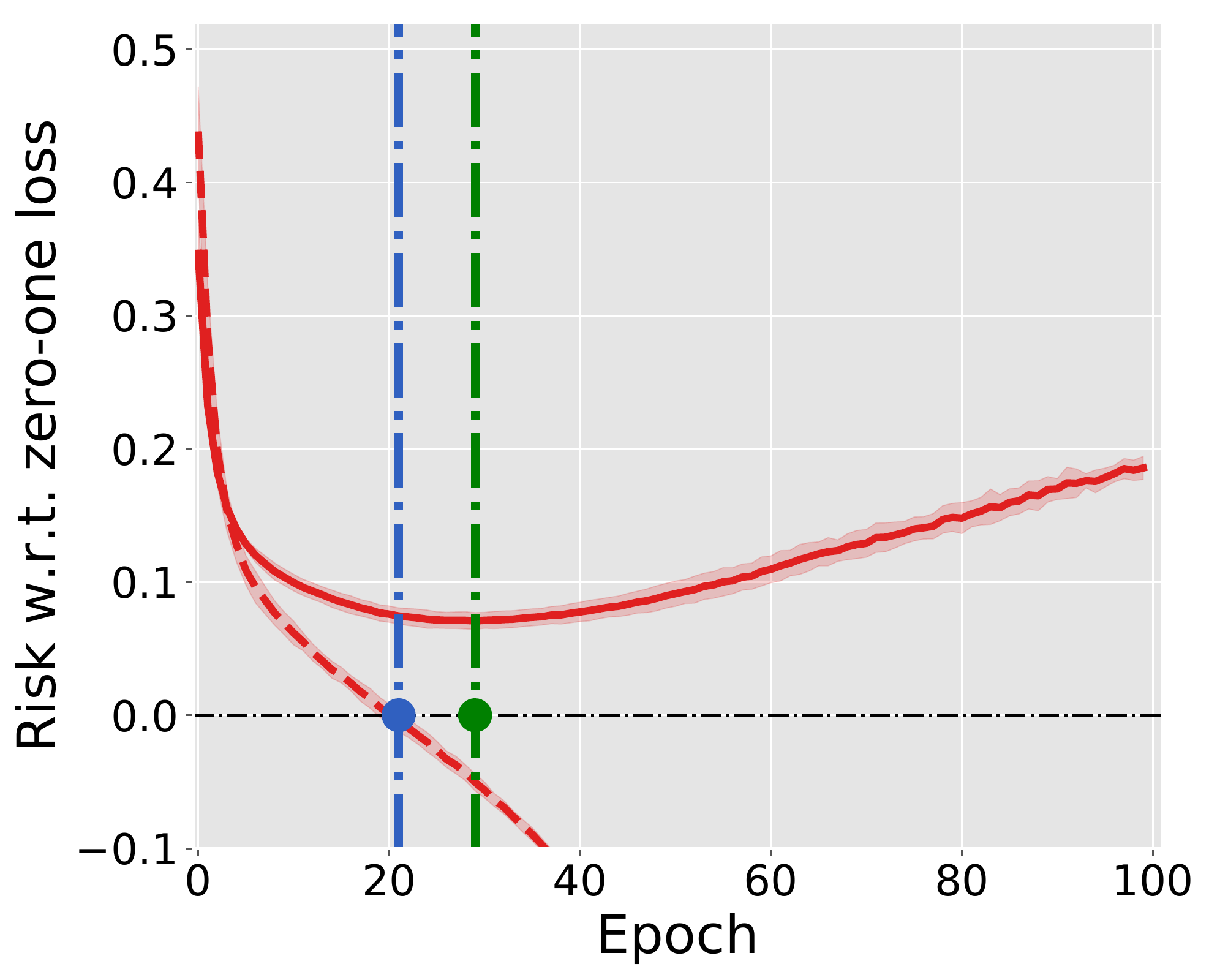}%
        \includegraphics[width=0.25\textwidth]{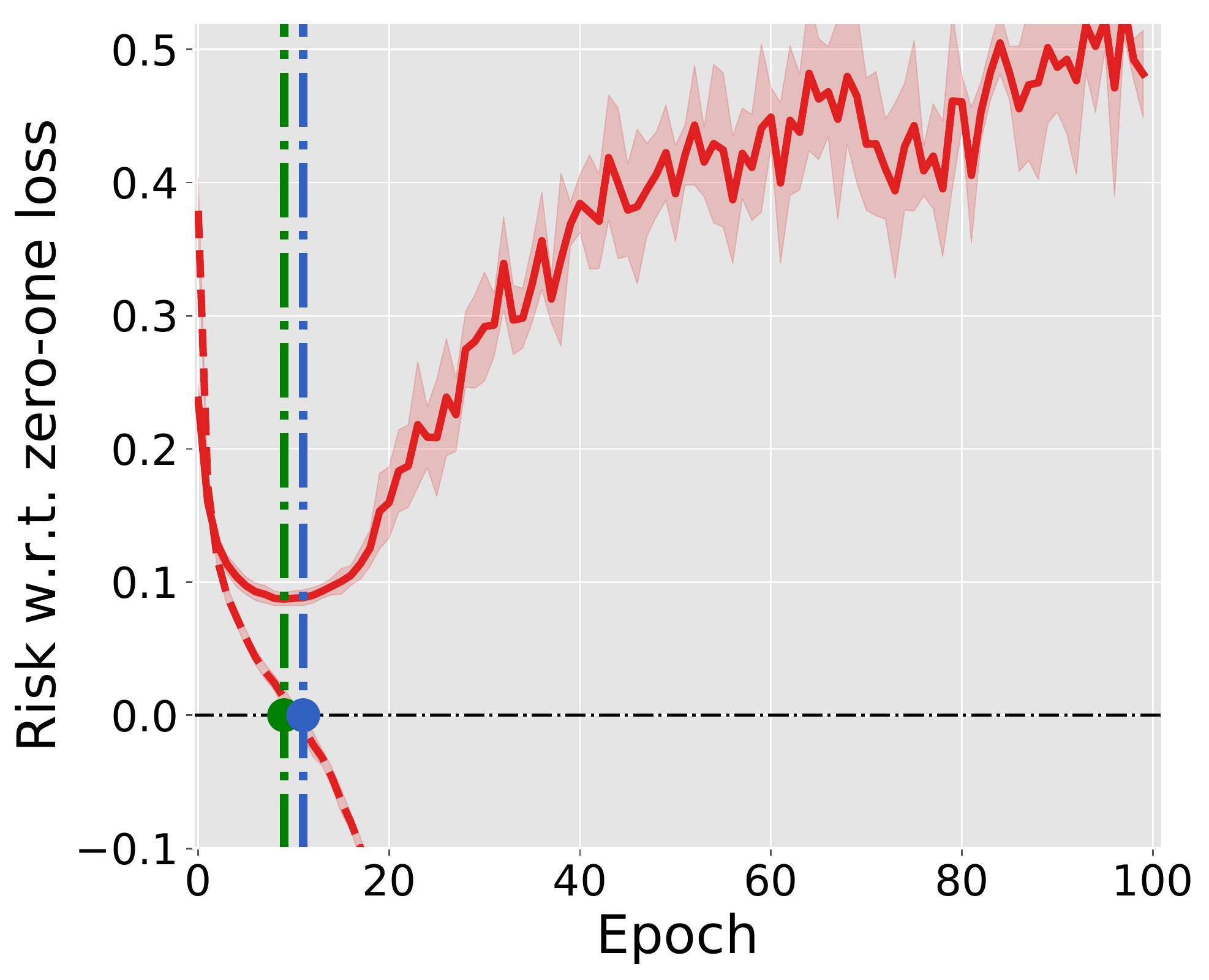}
    \end{minipage}\\
    \begin{minipage}[c]{0.20\textwidth}~\end{minipage}%
    \begin{minipage}[c]{0.2\textwidth}\centering\small AllConvNet, $\ellsig$ \end{minipage}%
    \begin{minipage}[c]{0.2\textwidth}\centering\small AllConvNet, $\elllog$ \end{minipage}%
    \begin{minipage}[c]{0.2\textwidth}\centering\small ResNet-32, $\ellsig$ \end{minipage}%
    \begin{minipage}[c]{0.2\textwidth}\centering\small ResNet-32, $\elllog$ \end{minipage}\\
    \begin{minipage}[c]{0.17\textwidth}\centering\small CIFAR-10 dataset\\Adam optimizer \end{minipage}%
    \begin{minipage}[c]{0.8\textwidth}
        \includegraphics[width=0.25\textwidth]{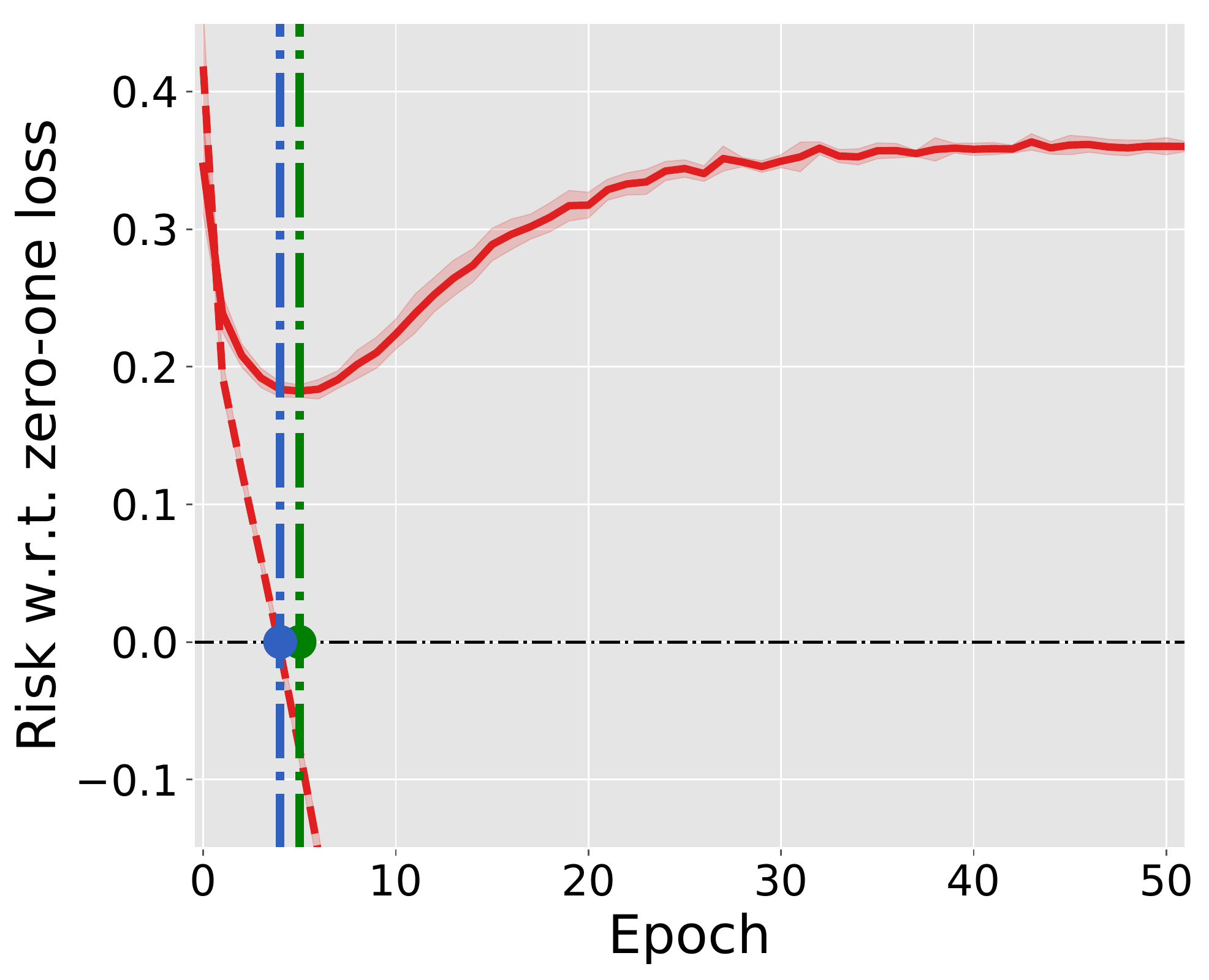}%
        \includegraphics[width=0.25\textwidth]{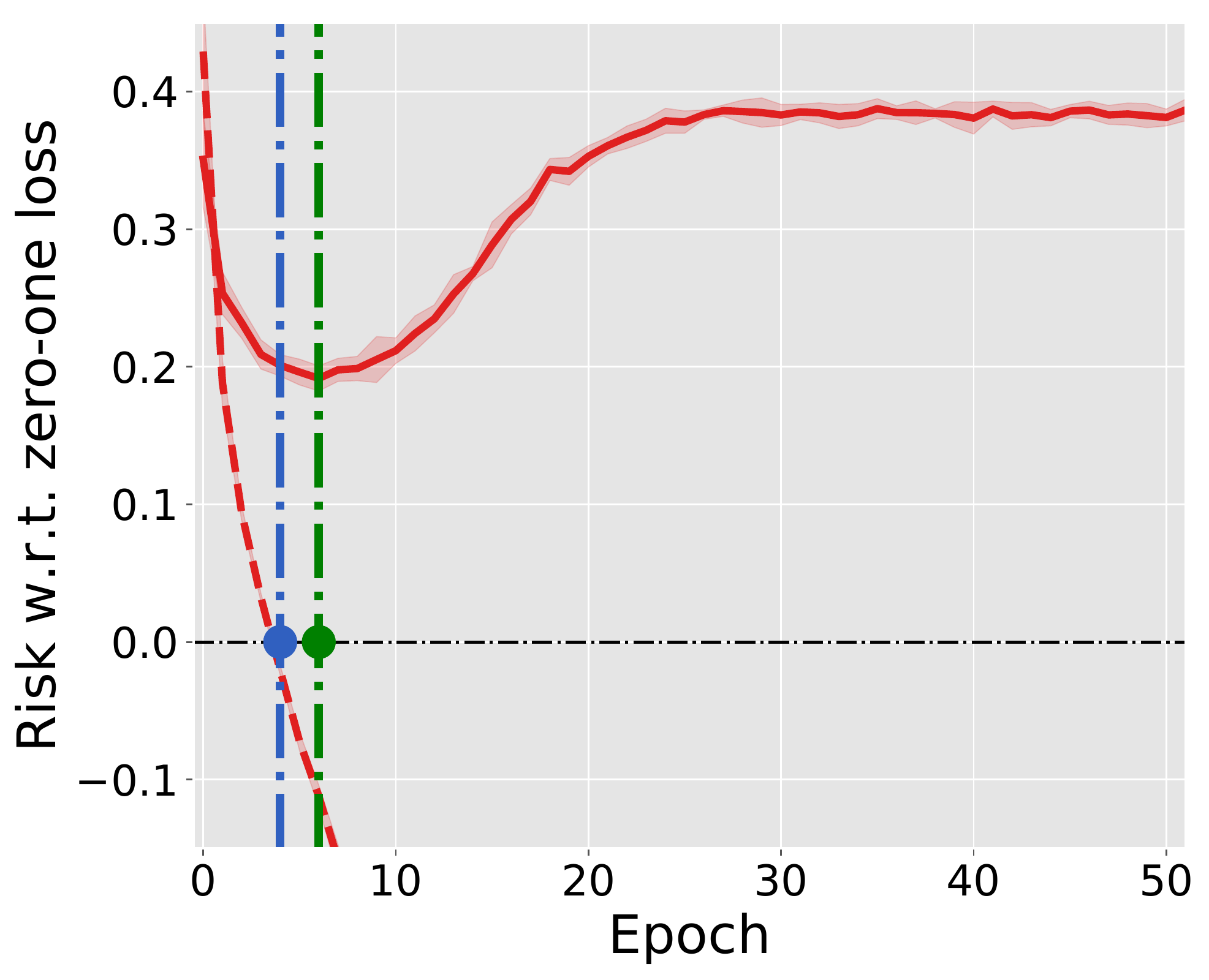}%
        \includegraphics[width=0.25\textwidth]{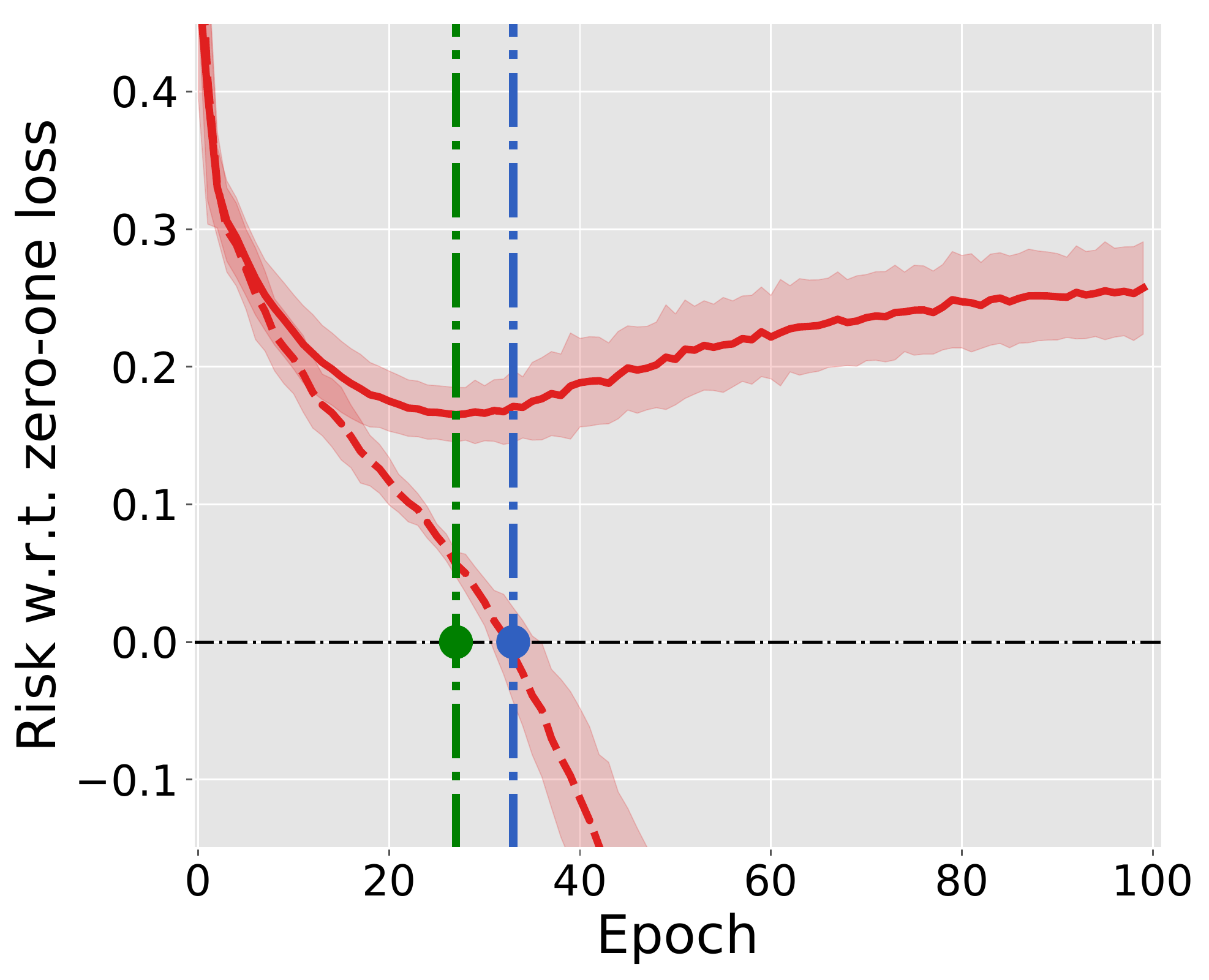}%
        \includegraphics[width=0.25\textwidth]{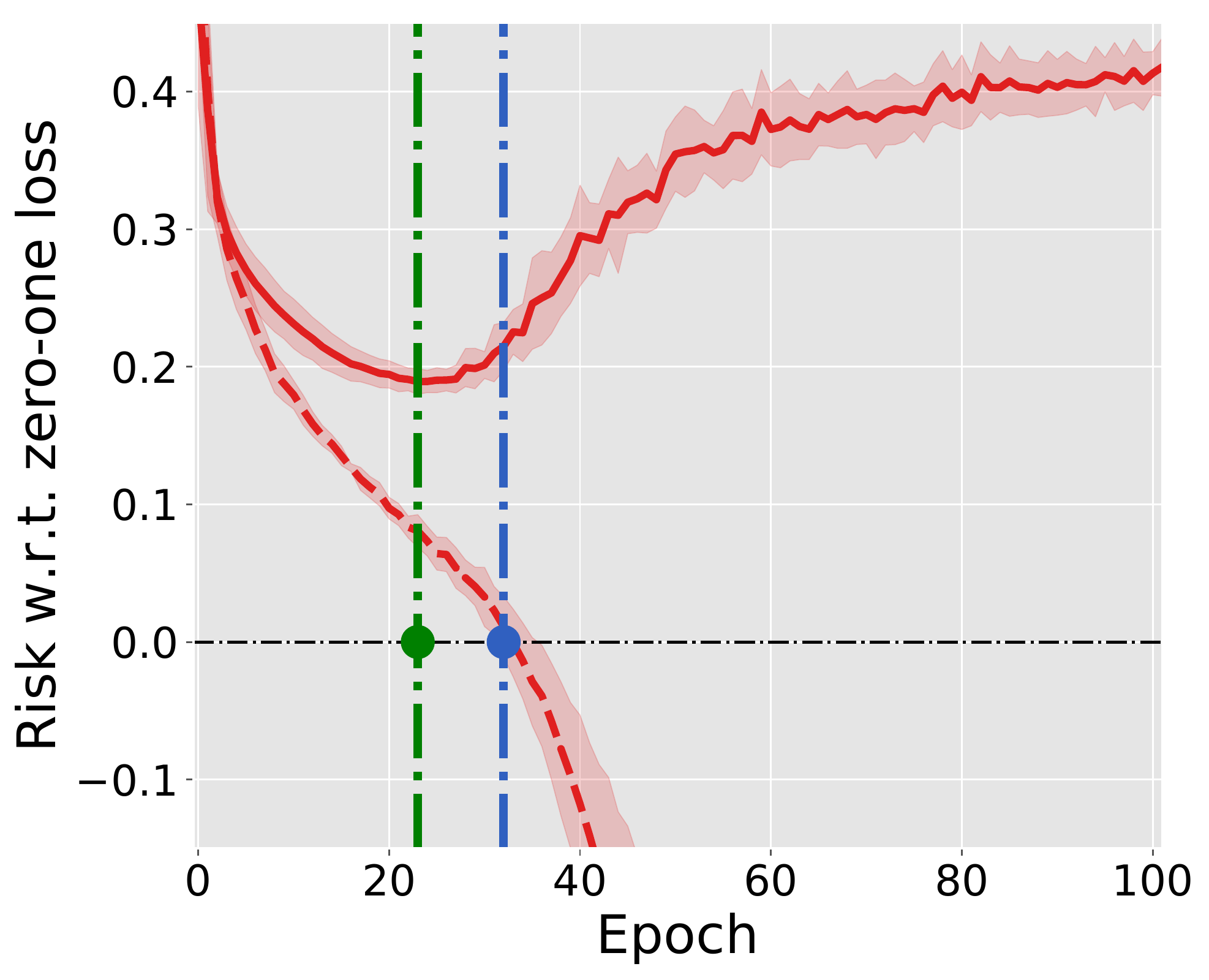}
    \end{minipage}\\
     \caption{Co-occurrence between severe overfitting and a negative empirical risk in UU classification. By co-occurrence, we mean both of them can be observed, or neither of them can be observed. In the upper row, on MNIST (even vs.~odd), a linear-in-input model~(Linear) and a 5-layer \emph{multi-layer perceptron}~(MLP) were trained by stochastic gradient descent~(SGD) using the sigmoid~($\ellsig$) and logistic~($\elllog$) losses. In the bottom row, on CIFAR-10 (transportation vs.~animal), the \emph{all convolutional net}~(AllConvNet) \citep{springenberg15iclr} and the 32-layer \emph{residual network}~(ResNet-32) \citep{he16cvpr} were trained by Adam \citep{kingma15iclr} using the same losses. The class priors $\theta$ and $\theta'$ were set to be $0.6$ and $0.4$ (see Sec.~\ref{sec:pre} for details). The blue dashed lines indicate when the empirical risk computed from UU training data goes negative; the green dashed lines indicate when the test error turns around and severe overfitting begins. We can clearly see a high co-occurrence in the figure regardless of datasets, optimizers, models, and losses. Details of how to reproduce the figure can be found in Appendix~\ref{sec:supp_figure1}.
      }
	\label{fig:illustration}
	\vspace{-2ex}%
\end{figure*}

\section{Introduction}
In traditional supervised classification, we always assume a vast amount of labeled data in the training phase. However, labeling industrial-level data can be expensive and time-consuming due to laborious mannual annotations. Furthermore, in some real-world problems such as medical diagnosis \citep{li2007improve,fakoor2013using,sun2017enhancing}, massive labeled data may not even be possible to collect. This has led to the development of machine learning algorithms to leverage large-scale unlabeled (U) data, including but not limited to semi-supervised learning \citep{grandvalet04nips,mann07icml,niu13icml,miyato16iclr,laine17iclr,luoi18cvpr,Oliver18nips} and positive-unlabeled learning \citep{elkan08kdd,christo14nips,christo15icml,niu16nips,kiryo17nips,kato19iclr}.

In this paper, we consider a more challenging setting of learning from only U data. A na\"{i}ve approach to this problem is to use \emph{discriminative clustering} \citep{xu04nips,valizadegan06nips,gomes10nips,sugiyama14neco,hu17icml}, which is also known as \emph{unsupervised classification}. But this solution is usually suboptimal due to the tacit \emph{clustering assumption} that \emph{one cluster corresponds to one class} \citep{chapelle02nips}, which is often violated in practice. For example, when one cluster is formed by a few geometrically close classes, or one class is formed by several geometrically separated clusters, even perfect clustering may still result in poor classification.


In order to avoid the unrealistic clustering assumption, we prefer to utilize U data for \emph{risk evaluation} and then optimize the obtained risk estimator by \emph{empirical risk minimization}~(ERM), as what has been carried out in standard supervised classification methods. This line of research was pioneered by \cite{christo13taai} and \cite{menon15icml}, where a binary classifier is trained from \emph{two sets of U data with different class priors}. However, a critical limitation is that their performance measure has to be the \emph{balanced error} \citep{brodersen10icpr}, which is the classification accuracy when the class prior is $1/2$. Recently, \cite{lu19iclr} extended these works and developed the first ERM method based on unbiased risk estimators for learning from two sets of U data. Their method, called \emph{UU classification} achieved the state-of-the-art performance in experiments. 

However, we found that, depending on the situation, the state-of-the-art unbiased UU method still suffers from severe overfitting as demonstrated in Figure~\ref{fig:illustration}. Based on our empirical explorations to this problem, we conjecture that the overfitting issue of the unbiased UU method is strongly connected to the empirical risk on training data going \emph{negative} due to the co-occurrence of them regardless of datasets, models, optimizers and loss functions. This negative empirical training risk should be fixed since the empirical training risk in standard supervised classification is always non-negative as long as the loss function is non-negative, which might be a potential reason for the unbiased risk estimator based method to overfit. \footnote{Note that the general-purpose regularization techniques, such as weight decay and dropout, fail to mitigate this overfitting as illustrated and analyzed in Appendix~\ref{sec:suppexp}.}

In this paper, we focus on mitigating this overfitting, where our goal is to learn a robust binary classifier from two U sets with different class priors following the ERM principle. To this end, we propose a novel \emph{consistent risk correction} technique that follows and improves the state-of-the-art unbiased UU method. The proposed method has the following advantages:

\begin{itemize}[leftmargin=*]
\item Empirically, the proposed corrected risk estimators are robust against overfitting. Theoretically, they are asymptotically unbiased and thus may be properly used for hyparameter tuning with only UU data, which is a clear advantage in deep learning since no labeled validation data are needed. Furthermore, the corrected minimizers possess an \emph{estimation error bound} which guarantees the \emph{consistency of learning} \citep{mohri12FML,sshwartz14UML};
\item We do not have implicit assumptions on the loss function, model architecture, and optimization, thus allowing the use of any loss (convex and non-convex), any model (e.g., the linear-in-parameter model and deep neural network) and any off-the-shelf \emph{stochastic optimization algorithms} \citep[e.g.,][]{duchi11jmlr,kingma15iclr}.
\end{itemize}
\paragraph{Organization} The rest of the paper is organized as follows. We formalize our research problem in Sec.~\ref{sec:pre}. In Sec.~\ref{sec:pratical_uu}, we propose the consistent risk correction with theoretical analysis. Experimental results are discussed in Sec.~\ref{sec:exp}, and conclusions are given in Sec.~\ref{sec:concl}. Proofs are presented in the supplementary material.
\vspace{-0.5ex}%
\section{Preliminaries}
\label{sec:pre}%
\vspace{-0.5ex}%
In this section, we introduce some notations and review the formulations of standard supervised classification and learning from two sets of U data with different class priors.
\vspace{-0.5ex}%
\subsection{Learning from fully labeled data}
\label{sec:fully_supervised}%
\vspace{-0.5ex}%
We begin with the standard supervised classification setup. Let $\mathcal{X}$ be the example space and $\mathcal{Y}=\{+1, -1\}$ be a binary label space. Denote by $\mathcal{D}$ the \emph{underlying joint distribution} over $\mathcal{X} \times \mathcal{Y}$. Any $\mathcal{D}$ may be decomposed into \emph{class-conditional distributions} $(P_\mathrm{p},P_\mathrm{n})=(p(x\mid y=+1), p(x\mid y=-1))$ and \emph{class-prior probability} $\pip=p(y=+1)$.

Let $g:\mathcal{X}\to\bR$ be an arbitrary binary classifier and $\ell:\bR\times\mathcal{Y}\to\bR_{+}$ be a \emph{loss function}, such that the value $\ell(t,y)$ means the loss for predicting the ground truth label $y$ by $t$. We assume that the loss function is non-negative. Denote by $\Rp^+(g)=\bE_{x\sim P_\mathrm{p}}[\ell(g(x),+1)]$ and $\Rn^-(g)=\bE_{x\sim P_\mathrm{n}}[\ell(g(x),-1)]$. The goal of binary classification is to obtain a classifier $g$ which minimizes the risk defined as
\begin{align}
\label{eq:risk}%
R(g) = \bE_{(x,y)\sim\mathcal{D}}[\ell(g(x),y)]= \pip\Rp^+(g)+\pin\Rn^-(g),
\end{align}
where $\bE_{(x,y)\sim \mathcal{D}}$ denotes the expectation over $\mathcal{D}$, and $\pin=p(y=-1)=1-\pip$. If $\ell$ is the \emph{zero-one loss} defined by $\ell_{01}(t,y)=(1-\sign(ty))/2$, the risk is named the \emph{classification error} (or the \emph{misclassification rate}), which is the standard performance measure in classification \citep{mohri12FML}. 

Since the joint distribution $\mathcal{D}$ is unknown, the ordinary ERM approach approximates the expectation by the average over training samples drawn i.i.d.~from $\mathcal{D}$ \citep{vapnik98SLT}. More specifically, given $\Xp=\{x^+_1,\ldots,x^+_\Np\}\stackrel{\mathrm{i.i.d.}}{\sim} P_\mathrm{p}$ and $\Xn=\{x^-_1,\ldots,x^-_\Nn\}\stackrel{\mathrm{i.i.d.}}{\sim} P_\mathrm{n}$, $R(g)$ can be approximated by \begin{align}
\label{eq:risk-pn-hat}%
\hRpn(g)=\pip\hRp^+(g)+\pin\hRn^-(g),
\end{align}
where $\hRp^+(g)=(1/\Np)\sum_{i=1}^{\Np}\ell(g(x^+_i),+1)$ and $\hRn^-(g)=(1/\Nn)\sum_{i=1}^{\Nn}\ell(g(x^-_i),-1)$.

\subsection{Learning from two sets of U data with different class priors}
\label{sec:unbiased_uu}%

Next we consider the problem of learning from two sets of U data with different class priors, which is called \emph{unlabeled-unlabeled}~(UU) \emph{classification} in \cite{lu19iclr}. We are given only unlabeled samples drawn from the following marginal distributions:
\begin{align}
    \ptr(x)&=\theta P_\mathrm{p}+(1-\theta) P_\mathrm{n},\notag\\
    \label{eq:train-density}
    \ptr'(x)&=\theta' P_\mathrm{p}+(1-\theta') P_\mathrm{n},
\end{align}
where $\theta$ and $\theta'$ are two class priors such that $\theta\neq\theta'$. This implies there are $\ptr(x,y)$ and $\ptr'(x,y)$, whose class-conditional densities are same and equal to those of $\mathcal{D}$, but whose class priors are different, i.e., 
\begin{align*}
    \ptr(x\mid y)&=\ptr'(x\mid y)=p(x\mid y),\\
    \ptr(y=+1)&=\theta\neq\theta'=\ptr'(y=+1).
\end{align*}
 More specifically, we have
 $\Xtr=\{x_1,\ldots,x_n\}\stackrel{\mathrm{i.i.d.}}{\sim}\ptr(x)$ and $\Xtr'=\{x'_1,\ldots,x'_{n'}\}\stackrel{\mathrm{i.i.d.}}{\sim}\ptr'(x)$, and our goal is to train a binary classifier that can generalize well with respect to the original $\mathcal{D}$.

In the standard supervised classification setting where training data are directly drawn from $\mathcal{D}$, the expectation in \eqref{eq:risk} can be estimated by the corresponding sample average. However, in the UU classification setting, no labeled samples are available and therefore the risk may not be estimated directly.

This problem can be avoided by the \emph{risk rewriting} approach \citep{lu19iclr,vanrooyen18jmlr}: the risk \eqref{eq:risk} is firstly rewritten into an equivalent expression such that it only involves the same distributions from which two sets of U data are sampled, and then it is estimated by plugging in the given U data. Let $R_{\mathrm{u}}^+(g)=\bE_{x\sim \ptr}[\ell(g(x),+1)]$, $R_{\mathrm{u}}^-(g)=\bE_{x\sim \ptr}[\ell(g(x),-1)]$, $R_{\mathrm{u'}}^+(g)=\bE_{x\sim \ptr'}[\ell(g(x'),+1)]$ and $R_{\mathrm{u'}}^-(g)=\bE_{x\sim \ptr'}\ell(g(x'),-1)]$. $R(g)$ can be expressed by
\begin{align*}
    \begin{split}
        R(g)&=aR_{\mathrm{u}}^+(g)-bR_{\mathrm{u}}^-(g)-cR_{\mathrm{u'}}^+(g)+dR_{\mathrm{u'}}^-(g),
    \end{split}
\end{align*}
where $a=\frac{(1-\theta')\pip}{\theta-\theta'}$, $b=\frac{\theta'(1-\pip)}{\theta-\theta'}$, $c=\frac{(1-\theta)\pip}{\theta-\theta'}$, and $d=\frac{\theta(1-\pip)}{\theta-\theta'}$.
Then with empirical estimates $\hRui^+(g)=\frac{1}{n}\sum_{i=1}^n\ell(g(x_i),+1)$, $\hRui^-(g)=\frac{1}{n}\sum_{i=1}^n\ell(g(x_i),-1)$, $\hRuii^+(g)=\frac{1}{n'}\sum_{j=1}^{n'}\ell(g(x'_j),+1)$, and $\hRuii^-(g)=\frac{1}{n'}\sum_{j=1}^{n'}\ell(g(x'_j),-1)$, $R(g)$ can be approximated as
\begin{align}
    \label{eq:risk-uu-hat}%
    \begin{split}
        \hRuu(g) &=a\hRui^+(g)-b\hRui^-(g)-c\hRuii^+(g)+d\hRuii^-(g).
    \end{split}
\end{align}
The \emph{empirical risk estimators} in Eqs.~\eqref{eq:risk-pn-hat} and \eqref{eq:risk-uu-hat} are \emph{unbiased} and \emph{consistent}%
\footnote{The consistency here means for fixed $g$, $\hRpn(g)\to R(g)$ and $\hRuu(g)\to R(g)$ as $\Np,\Nn,n,n'\to\infty$.} w.r.t.~all loss functions. When they are used for evaluating the classification accuracy, $\ell$ is by default $\ell_{01}$; when they are used for training, it is replaced with a \emph{surrogate loss} since $\ell_{01}$ is discontinuous and therefore difficult to optimize \citep{bendavid06jcss,bartlett06jasa}.

The unbiased risk estimator methods \citep{lu19iclr,vanrooyen18jmlr} use classification error \eqref{eq:risk} as the performance measure and assume the knowledge of class priors. Note that given only U data, by no means could we learn the class priors without any assumptions \citep{menon15icml}. But, by introducing the \emph{mutually irreducible condition} \citep{scott13colt}, the class priors become identifiable and can be estimated in some cases \citep{menon15icml,liu16tpami,jain2016estimating,blanchard2016}. To simplify analysis, we assume the class priors to be known in this paper.

Another line of research on UU classification focuses on the \emph{balanced error}~(BER), which is a special case of the classification error \eqref{eq:risk}, defined by $B(g) = \frac{1}{2}\bE_{x\sim P_\mathrm{p}}[\ell_{01}(g(x),+1)]\!+\!\frac{1}{2}\bE_{x\sim P_\mathrm{n}}[\ell_{01}(g(x),-1)]$. Though BER minimization methods do not need the knowledge of class priors, they assume that the class prior is balanced (i.e., $\pip=\frac{1}{2}$) \citep{christo13taai,menon15icml,Nutt19icml}. Note that $B(g) =R(g)$ for any $g$ if and only if $\pip=\frac{1}{2}$, which indicates that BER is a meaningful performance measure for classification when $\pip\approx\frac{1}{2}$ while it definitely biases learning when $\pip\approx\frac{1}{2}$ is not the case. Therefore, through out this paper, we consider the more natural classification error metric \eqref{eq:risk}.

\section{Consistent Risk Correction}
\label{sec:pratical_uu}%

In this section, we first study the overfitting issue of the unbiased risk estimator of UU classification, and then propose our consistent risk correction method with theoretical guarantees.

\subsection{Is an unbiased risk estimator really good?}
\label{sec:unbiased}%

As discussed in Sec.~\ref{sec:unbiased_uu}, the state-of-the-art method of UU classification uses the risk rewriting technique to obtain an unbiased risk estimator. However, the derived unbiased UU risk estimator \eqref{eq:risk-uu-hat} contains two negative partial risks $-b\hRui^-(g)$ and $-c\hRuii^+(g)$. This may be problematic since the original expression of the classification risk \eqref{eq:risk} only includes expectations over non-negative loss $\ell:\bR\times\mathcal{Y}\to\bR_{+}$ and is by definition non-negative. In practice, we find that the unbiased UU method may suffer severe overfitting and observe a high co-occurrence between overfitting and their empirical risk going negative. Thus, we conjecture that the negative empirical risk might be a potential reason that results in overfitting.

We elaborate on the issue in Figure~\ref{fig:illustration}, where we trained various models on MNIST and CIFAR-10 using different optimizers and loss functions. From the experimental results, we can see a strong co-occurrence of severe overfitting and a negative empirical risk regardless of datasets, models, optimizers, and loss functions: in the experiments of MNIST and CIFAR-10 with different deep neural network models, optimizers, and loss functions, overfitting is observable when the empirical risk on the training data goes negative; in the experiments on MNIST with the linear model and SGD optimizer, the test performance is reasonably good while the empirical risk on the training data is kept non-negative. The overfitting is more severe when flexible models such as deep neural networks are used, since they have larger capacity to fit data and thus they make the negative partial risks $-b\hRui^-(g)$ and $-c\hRuii^+(g)$ more negative.

\subsection{Corrected risk estimator}
\label{sec:nnuu}%
Now we face a dilemma: in many real-world problems, we may only collect large unlabeled datasets and still wish our classifier trained from them generalize well. So the question arises: can we alleviate the aforementioned overfitting problem with neither labeling more training data nor turning to a suboptimal solution (e.g., clustering)?

The answer is affirmative. In Figure~\ref{fig:illustration}, we observed that the resulting empirical risk $\hRuu(g)$ keeps decreasing and goes negative. This issue should be fixed since the empirical training risk in standard supervised classification is always non-negative for non-negative loss functions. Note that the two terms (i.e., $\Rp^+(g)$ and $\Rn^-(g)$) in the original classification risk \eqref{eq:risk}, which correspond to the risks of the P and N classes, are both non-negative. Thus our basic idea is reformulating the rewritten risk \eqref{eq:risk-uu-hat} to find the counterparts for the risks of the P and N classes in \eqref{eq:risk}:
\begin{align*}
    \pip\Rp^+(g)&=aR_{\mathrm{u}}^+(g)-cR_{\mathrm{u'}}^+(g),\\
    \pin\Rn^-(g)&=dR_{\mathrm{u'}}^-(g)-bR_{\mathrm{u}}^-(g).
\end{align*}
We then enforce non-negativity to these counterparts. More specifically, we have
\begin{align}
\widehat{R}_{\mathrm{uu}\text{-}\mathrm{max}}(g) &=\max\left\{0,a\hRui^+(g)-c\hRuii^+(g)\right\}\notag\\
\label{eq:risk-uu-max}%
&\quad +\max\left\{0,d\hRuii^-(g)-b\hRui^-(g)\right\}.
\end{align}

This is motivated by \cite{kiryo17nips}, which considered the problem of the rewritten risk going negative in the context of positive-unlabeled learning. In their setting, the reformulated P risk is exactly the same as its counterpart in the original classification risk \eqref{eq:risk} (i.e., $\Rp^+(g)$), since they are given positive data with true labels. So there was only one max operator in the reformulated N risk. Our setting differs from them since we are given only unlabeled data and therefore needs the ``max'' correction for both the reformulated P and N risks.

\begin{figure}[t]
    \centering
    \includegraphics[width=0.35\textwidth]{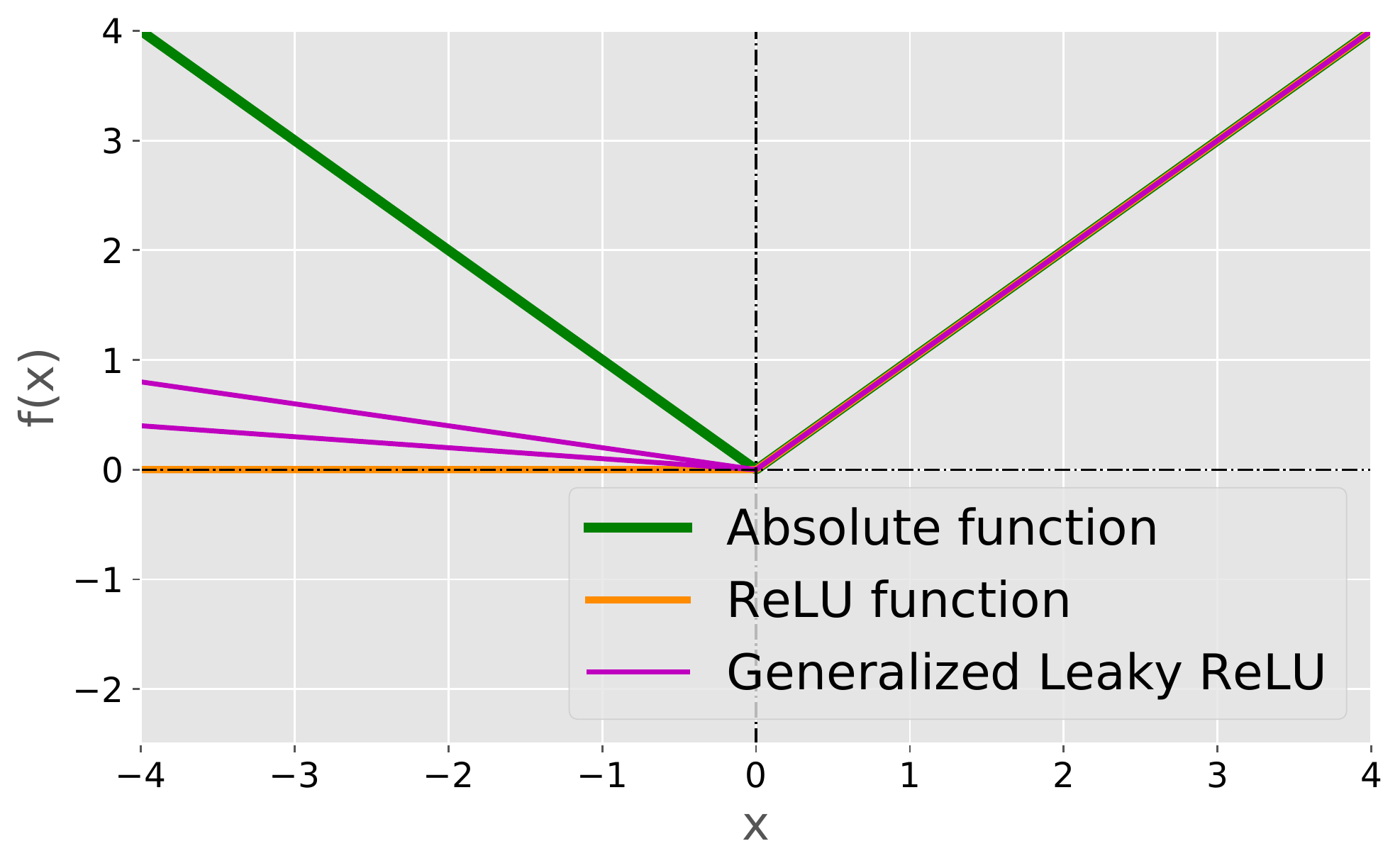}
    \caption{Examples of consistent correction functions.}
    \label{fig:examples}
\end{figure}

However, the max operator completely ignores the training data that yield a negative risk. We argue that the information in those data is also useful for training and should not be dropped. Following this idea, we propose a generalized \emph{consistent correction function} as follows:
\begin{definition}[Consistent correction function]
\label{def:cc}%
A function $f:\bR\to\bR$ is called a consistent correction function if it is Lipschitz continuous, non-negative and $f(x)=x$ for all $x\ge0$. Let $\cF$ be a class of all consistent correction functions.
\end{definition}

For example, the rectified linear unit~(ReLU) function and absolute value function belong to $\cF$. Based on this definition, we propose a family of consistently corrected risk estimators $\hRnnuu$ by
\begin{align}
\hRnnuu(g) &=\textstyle f_1\left(a\hRui^+(g)-c\hRuii^+(g)\right)\notag\\
\label{eq:nn-emp-risk-uu-hat}%
&\quad +f_2\left(d\hRuii^-(g)-b\hRui^-(g)\right),
\end{align}
where $f_1$ and $f_2$ can be any consistent correction funtions. The proposed corrected risk estimator is by nature ERM-based, and consequently the empirical risk minimizer of \eqref{eq:nn-emp-risk-uu-hat}, i.e., $\hgcc=\argmin_{g\in\cG}\hRnnuu(g)$ can be obtained by flexible models and powerful stochastic optimization algorithms.

The corrected UU classification algorithm is described in Algorithm~\ref{alg:large-scale-uu}. In the implementation, we propose to use the generalized leaky ReLU function, i.e., $f(x)=f_1(x)=f_2(x)=\mathbb{I}_{\{x\geq 0\}}x+\mathbb{I}_{\{x<0\}}\lambda x$, where $\lambda\leq 0$. The intuition behind is that instead of completely ignoring the training data that yield a negative risk by the ``max'' correction, we propose to actively control learning on those sensitive data by adding weights on the negative partial risks. Note that the ReLU function and the absolute value function are special cases of the generalized leaky ReLU function as illustrated in Figure~\ref{fig:examples}.%
\footnote{Using the ReLU and absolute value function to prevent the negative risk problem has been studied in the context of positive-unlabeled learning \citep{kiryo17nips} and complementary-label learning \citep{ishida19iclr}. Our proposal can be regarded as their extension to a family of correction functions applied in the UU classification setting.} 

\begin{algorithm}[t]
  \caption{Corrected UU classification}
  \label{alg:large-scale-uu}
  \begin{algorithmic}
    \STATE \textbf{Input:} two sets of U training data $(\Xtr,\Xtr')$
    \STATE \textbf{Output:} learned model parameter $\theta$
  \end{algorithmic}
  \begin{algorithmic}[1]
    \STATE Initialize $\theta$
    \STATE Let $\cA$ be an SGD-like optimizer working on $\theta$
    \STATE \textbf{for} $t=1$ \textbf{to} number\_of\_epochs:
    \STATE \quad Shuffle $(\Xtr,\Xtr')$
    \STATE \quad \textbf{for} $i=1$ \textbf{to} number\_of\_mini-batches:
    \STATE \qquad Let $(\bXtr,\bXtr')$ be the current mini-batch
    \STATE \qquad Forward $\bXtr$ and $\bXtr'$
    \STATE \qquad Compute\\
            $L^+ = a\ell(g(\bXtr),+1)/|\bXtr| -c\ell(g(\bXtr'),+1)/|\bXtr'|$\\
            $L^- = d\ell(g(\bXtr'),-1)/|\bXtr'| -b\ell(g(\bXtr),-1)/|\bXtr|$
    \STATE \qquad Correct them by $L^+_\mathrm{cc} = f(L^+),L^-_\mathrm{cc} = f(L^-)$
    \STATE \qquad Backward $L_\mathrm{cc} = L^+_\mathrm{cc}+L^-_\mathrm{cc}$
    \STATE \qquad Update $\theta$ by $\cA$
  \end{algorithmic}
\end{algorithm}

\begin{table*}[ht]
    \caption{Specification of benchmark datasets and models.}
    \label{tab:dataset}
    \vspace{-1ex}%
    \begin{center}\small
        \begin{tabular*}{0.8\textwidth}{c|cccc|cc}
            \toprule
            Dataset & \# Train & \# Test & \# Feature & $\pip$ & Simple $g(x)$ & Deep $g(x)$ \\
            \midrule
            MNIST & 60,000 & 10,000 & 784 & 0.50 & Linear model & 5-layer MLP \\
            Fashion-MNIST & 60,000 & 10,000 & 784 & 0.40 & Linear model & 5-layer MLP \\
            Kuzushiji-MNIST & 60,000 & 10,000 & 784 & 0.30 & Linear model & 5-layer MLP \\
            CIFAR-10 & 50,000 & 10,000 & 3,072 & 0.60 & Linear model & ResNet-32 \\
            \bottomrule
         \end{tabular*}
    \end{center}
    \vspace{-1em}%
\end{table*}

\subsection{Theoretical analysis}
\label{sec:theory}%
In this section, we analyze the consistently corrected risk estimator \eqref{eq:nn-emp-risk-uu-hat} and its minimizer.
\paragraph{Bias and consistency}
The proposed corrected risk estimator $\hRnnuu(g)$ is no longer unbiased due to the fact that $\hRuu(g)$ is unbiased and $\hRnnuu(g)\geq\hRuu(g)$ for any $(\Xtr,\Xtr')$ if we fix $g$. The question then arises: is $\hRnnuu(g)$ consistent? Next we prove the consistency. 

First, let $A=a\hRui^+(g)$, $B=b\hRui^-(g)$, $C=c\hRuii^+(g)$, $D=d\hRuii^-(g)$ and $L_f$ be the Lipschitz constant of $f_1$ and $f_2$. Then partition all possible $(\Xtr,\Xtr')$ into: $\fD^+(g) = \{(\Xtr,\Xtr') \mid A-C\ge0, D-B\ge0 \}$, $\fD^-(g) = \{(\Xtr,\Xtr') \mid A-C<0\}\cup\{(\Xtr,\Xtr') \mid D-B<0 \}$. Assume there are $C_g>0$ and $C_\ell>0$ such that $\sup_{g\in\cG}\|g\|_\infty\le C_g$ and $\sup_{|z|\le C_g}\ell(z)\le C_\ell$. By \emph{McDiarmid's inequality} \citep{mcdiarmid89MBD}, we can prove the following lemma.

\begin{lemma}
  \label{thm:pr-diff}%
  The bias of $\hRnnuu(g)$ is positive if and only if the probability measure of $\fD^-(g)$\footnote{The probability measure is induced by the randomness of the two unlabeled datasets, see Appendix~\ref{subsec:Lemma2} for the formal definition.} is non-zero.
  Further, by assuming that there is $\alpha_{g}>0$ and $\beta_{g}>0$ such that $\Rp^+(g)\ge\alpha_{g}/\pip$ and $\Rn^-(g)\ge\beta_{g}/\pin$, the probability measure of $\fD^-(g)$ can be bounded by
\begin{align}
\pr(\fD^-(g)) &\le \exp\left(-\frac{2\alpha_{g}^2/C_\ell^2}{a^2/n+c^2/n'}\right)\notag\\
  \label{eq:pr-diff-bound}%
&\quad +\exp\left(-\frac{2\beta_{g}^2/C_\ell^2}{b^2/n'+d^2/n}\right).
\end{align}
\end{lemma}

Based on Lemma~\ref{thm:pr-diff}, we can show the exponential decay of the bias and also the consistency.

\begin{theorem}[Bias and consistency]
  \label{thm:bias-consistency}%
  Let $\Delta_{g}=\exp\left(-\frac{2\alpha_{g}^2/C_\ell^2}{a^2/n+c^2/n'}\right)+\exp\left(-\frac{2\beta_{g}^2/C_\ell^2}{b^2/n'+d^2/n}\right)$. By assumption in Lemma~\ref{thm:pr-diff}, the bias of $\hRnnuu(g)$ decays exponentially as $n,n'\to\infty$:
  \begin{align}
  0 &\le \bE_{\Xtr,\Xtr'}[\hRnnuu(g)]-R(g)\notag\\
  \label{eq:bias-bound}%
  &\quad \le (L_f+1)(a+b+c+d)C_\ell\Delta_{g}.
  \end{align}
  Moreover, for any $\delta>0$, let $C_\delta=C_\ell L_f\sqrt{\ln(2/\delta)/2}$, $\chi_{n,n'}=(a+b)/\sqrt{n}+(c+d)/\sqrt{n'}$, then we have with probability at least $1-\delta$,
  \begin{align}
  \label{eq:dev-bound}%
  |\hRnnuu(g)-R(g)| &\le (L_f+1)(a+b+c+d)C_\ell\Delta_{g}\notag\\
  &\quad +C_\delta\cdot\chi_{n,n'},
  \end{align}
  and with probability at least $1-\delta-\Delta_{g}$,
  \begin{align}
  \label{eq:dev-bound-alter}%
  |\hRnnuu(g)-R(g)| \le C_\delta\cdot\chi_{n,n'}.
  \end{align}
\end{theorem}

Either \eqref{eq:dev-bound} or \eqref{eq:dev-bound-alter} in Theorem~\ref{thm:bias-consistency} indicates for fixed $g$, $\hRnnuu(g)\to R(g)$ in $\cO_p(1/\sqrt{n}+1/\sqrt{n'})$. This convergence rate is optimal according to the \emph{central limit theorem} \citep{chung68CPT}, which means the proposed estimator is a biased yet optimal estimator to the risk.
\paragraph{Estimation error bound}
While Theorem~\ref{thm:bias-consistency} addressed the use of \eqref{eq:nn-emp-risk-uu-hat} when the risk is evaluated, in what follows we study the estimation error $R(\hgcc) -R(g^*)$ when classifiers are trained, where $g^*$ is the true risk minimizer in the model class $\cG$, i.e., $g^*=\argmin_{g\in\cG}R(g)$. As a common practice \citep{mohri12FML,boucheron2005theory}, assume that the instances are upper bounded, i.e., $\|x\|\le C_x$, and that the loss function $\ell(t,y)$ is Lipschitz continuous in $t$ for all $|t|\le C_g$ with a Lipschitz constant $L_\ell$.

\begin{theorem}[Estimation error bound]
  \label{thm:est-err}%
  Assume that
  (a) $\inf_{g\in\cG}\Rp^+(g)\ge\alpha/\pip>0$, $\inf_{g\in\cG}\Rn^-(g)\ge\beta/\pin>0$;
  (b) $\cG$ is closed under negation, i.e., $g\in\cG$ if and only if $-g\in\cG$.
  Let $\Delta=\exp\left(-\frac{2\alpha^2/C_\ell^2}{a^2/n+c^2/n'}\right)+\exp\left(-\frac{2\beta^2/C_\ell^2}{b^2/n'+d^2/n}\right)$. Then, for any $\delta>0$, with probability at least $1-\delta$,
  \begin{align}
  &R(\hgcc)-R(g^*)\notag\\
  &\le 8(a+b)L_fL_\ell\fR_{n,\ptr}(\cG)+8(c+d)L_fL_\ell\fR_{n',\ptr'}(\cG)\notag\\
  \label{eq:est-err-bound}%
  &\quad +2(L_f+1)(a+b+c+d)C_\ell\Delta+2C_\delta'\cdot\chi_{n,n'},
  \end{align}
  where $C_\delta'=C_\ell L_f\sqrt{\ln(1/\delta)/2}$, and $\fR_{n,\ptr}(\cG)$ and $\fR_{n',\ptr'}(\cG)$ are the Rademacher complexities of $\cG$ for the sampling of size $n$ from $\ptr(x)$ and of size $n'$ from $\ptr'(x)$, respectively.
\end{theorem}

Theorem~\ref{thm:est-err} ensures that learning with \eqref{eq:nn-emp-risk-uu-hat} is also consistent: as $n,n'\to\infty$, $R(\hgcc)\to R(g^*)$, since $\fR_{n,\ptr}(\cG)$, $\fR_{n',\ptr'}(\cG)\to0$ for all parametric models with a bounded norm and $\Delta\to0$. Specifically, for linear-in-parameter models with a bounded norm, $\fR_{n,\ptr}(\cG)=\cO(1/\sqrt{n})$ and $\fR_{n',\ptr'}(\cG)=\cO(1/\sqrt{n'})$, and thus $R(\hgcc)\to R(g^*)$ in $\cO_p(1/\sqrt{n}+1/\sqrt{n'})$. Furthermore, for deep neural networks, we can obtain the following corollary based on the results in \cite{golowich2017size}.

Consider neural networks of the form $g:x\mapsto W_m\sigma_{m-1}(W_{m-1}\sigma_{m-2}(\ldots\sigma_1(W_1 x)))$, where $m$ is the depth of the neural network, $W_1,\ldots,W_m$ are weight matrices, and $\sigma_1,\ldots,\sigma_{m-1}$ are activation functions for each layer.

\begin{corollary}
  \label{thm:deep-bound}%
  Assume the Frobenius norm of the weight matrices $W_j$ are at most $M_{F}(j)$. Let $\sigma$ be a positive-homogeneous (i.e., it is element-wise and satisfies $\sigma(\alpha z)=\alpha\sigma(z)$ for all $\alpha\ge 0$ and $z\in\bR$), $1$-Lipschitz activation function which is applied element-wise (such as the ReLU). Then, for any $\delta>0$, with probability at least $1-\delta$,
\begin{align*}
&R(\hgcc)-R(g^*)\le\\
&\left(8L_f L_\ell C_x\left(\sqrt{2m\log 2}+1\right)\prod_{j=1}^{m} M_{F}(j)+2C_\delta'\right)\cdot\chi_{n,n'}\\
&+2(L_f+1)(a+b+c+d)C_\ell\Delta.
\end{align*}
\end{corollary}

The factor $(\sqrt{2m\log 2}+1) \prod_{j=1}^{m} M_{F}(j)$ is induced by the hypothesis complexity of the deep neural network and could be improved \citep{golowich2017size}. From  Corollary~\ref{thm:deep-bound}, for fully connected neural networks, we obtain the same convergence rate as the linear-in-parameter models. 

\begin{table*}[t]
    \caption{Means (standard deviations) of the classification accuracy (Acc) and the drop ($\Delta_A$) over five trials in percentage with simple models. The best and comparable methods based on the paired \textit{t}-test at the significance level 5\% are highlighted in boldface.}
    \label{tab:simple}
    \begin{center}\scriptsize
    \resizebox{170mm}{!}{
    \begin{tabular*}{0.855\textwidth}{cccccccccccc}
            \toprule
            \multirow{2}{*}{Dataset} & \multirow{2}{*}{$\theta$, $\theta'$} & \multicolumn{2}{c}{UU-Biased} & \multicolumn{2}{c}{UU-Unbiased} & \multicolumn{2}{c}{UU-ABS} & \multicolumn{2}{c}{UU-ReLU} & \multicolumn{2}{c}{UU-LReLU}\\
            \cmidrule(r){3-4} \cmidrule(r){5-6} \cmidrule(r){7-8} \cmidrule(r){9-10} \cmidrule(r){11-12}
            & & Acc & $\Delta_{A}$ & Acc & $\Delta_{A}$ & Acc & $\Delta_{A}$ & Acc & $\Delta_{A}$ & Acc & $\Delta_{A}$\\
            \midrule
            
            \multirow{6}{*}{MNIST} & \multirow{2}{*}{0.8, 0.2} & 89.30 & 0.28 & \bf89.76 & 0.10 & \bf89.80 & 0.10 & \bf89.68 & 0.14 & \bf89.70 & 0.12\\
            & & (0.09) & (0.12) & \bf(0.13) & (0.04) & \bf(0.20) & (0.03) & \bf(0.14) & (0.07) & \bf(0.14) & (0.07)\\
            & \multirow{2}{*}{0.7, 0.3} & 88.59 & 0.42 & \bf89.26 & 0.11 & \bf89.19 & 0.21 & \bf89.24 & 0.14 & \bf89.15 & 0.20\\
            & & (0.22) & (0.11) & \bf(0.09) & (0.07) & \bf(0.20) & (0.06) & \bf(0.11) & (0.07) & \bf(0.27) & (0.06)\\
            & \multirow{2}{*}{0.6, 0.4} & 84.64 & 2.03 & \bf87.15 & 0.54 & \bf87.28 & 0.49 & \bf87.13 & 0.60 & \bf87.26 & 0.40\\
            & & (0.42) & (0.15) & \bf(0.34) & (0.22) & \bf(0.38) & (0.23) & \bf(0.33) & (0.25) & \bf(0.37) & (0.08)\\
            \midrule
            
            \multirow{6}{*}{\tabincell{c}{Fashion-\\MNIST}} & \multirow{2}{*}{0.8, 0.2} & \bf87.27 & 0.82 & \bf87.73 & 0.43 & \bf87.72 & 0.44 & \bf87.78 & 0.35 & \bf87.78 & 0.39\\
            & & \bf(0.83) & (0.73) & \bf(0.11) & (0.06) & \bf(0.11) & (0.06) & \bf(0.20) & (0.14) & \bf(0.20) & (0.13)\\
            & \multirow{2}{*}{0.7, 0.3} & 85.53 & 2.02 & \bf86.99 & 0.73 & \bf87.02 & 0.71 & \bf87.07 & 0.72 & \bf86.84 & 0.97\\
            & & (0.93) & (0.77) & \bf(0.17) & (0.14) & \bf(0.35) & (0.26) & \bf(0.28) & (0.20) & \bf(0.56) & (0.52)\\
            & \multirow{2}{*}{0.6, 0.4} & 80.66 & 4.59 & \bf83.69 & 2.70 & \bf84.18 & 2.41 & \bf84.20 & 2.08 & \bf83.92 & 2.59\\
            & & (2.22) & (1.91) & \bf(0.53) & (0.46) & \bf(0.57) & (0.57) & \bf(0.44) & (0.63) & \bf(1.07) & (1.09)\\
            \midrule
            
            \multirow{6}{*}{\tabincell{c}{Kuzushiji-\\MNIST}} & \multirow{2}{*}{0.8, 0.2} & 72.73 & 1.59 & \bf79.19 & 0.39 & \bf79.28 & 0.39 & \bf79.28 & 0.39 & \bf79.32 & 0.35\\
            & & (0.39) & (0.45) & \bf(0.29) & (0.18) & \bf(0.29) & (0.18) & \bf(0.29) & (0.18) & \bf(0.19) & (0.20)\\
            & \multirow{2}{*}{0.7, 0.3} & 72.21 & 1.91 & \bf78.67 & 0.75 & \bf78.89 & 0.75 & \bf78.79 & 0.63 & \bf78.90 & 0.63\\
            & & (0.74) & (0.52) & \bf(0.34) & (0.25) & \bf(0.40) & (0.23) & \bf(0.21) & (0.25) & \bf(0.40) & (0.28)\\
            & \multirow{2}{*}{0.6, 0.4} & 69.76 & 3.14 & \bf77.73 & 1.24 & \bf77.95 & 1.20 & \bf77.84 & 1.26 & \bf77.86 & 1.19\\
            & & (0.46) & (0.70) & \bf(0.37) & (0.24) & \bf(0.71) & (0.43) & \bf(0.65) & (0.36) & \bf(0.72) & (0.29)\\
            \midrule
            
            \multirow{6}{*}{\tabincell{c}{CIFAR-\\10}} & \multirow{2}{*}{0.8, 0.2} & 76.94 & 4.62 & \bf80.50 & 1.49 & \bf80.48 & 1.50 & \bf80.82 & 1.07 & \bf81.13 & 0.76\\
            & & (5.49) & (5.35) & \bf(1.20) & (1.22) & \bf(1.19) & (1.21) & \bf(0.69) & (0.58) & \bf(0.51) & (0.41)\\
            & \multirow{2}{*}{0.7, 0.3} & \bf78.04 & 2.22 & \bf79.68 & 1.54 & \bf80.12 & 1.20 & \bf80.28 & 1.03 & \bf79.95 & 1.32\\
            & & \bf(2.02) & (2.18) & \bf(0.66) & (0.56) & \bf(0.42) & (0.35) & \bf(0.14) & (0.21) & \bf(0.67) & (0.59)\\
            & \multirow{2}{*}{0.6, 0.4} & 67.23 & 9.05 & \bf76.34 & 3.74 & \bf75.21 & 4.81 & \bf76.24 & 3.85 & \bf76.28 & 3.72\\
            & & (6.68) & (6.77) & \bf(1.41) & (1.51) & \bf(1.95) & (1.93) & \bf(0.96) & (0.99) & \bf(0.92) & (1.06)\\
            
            \bottomrule
        \end{tabular*}}
    \end{center}
\end{table*}

\section{Experiments}
\label{sec:exp}%

In this section, we verify the effectiveness of the proposed consistent risk correction methods on various models and datasets, and test under different class prior settings for an extensive investigation. 

\paragraph{Datasets}
We train on widely adopted benchmarks MNIST, Fashion-MNIST, Kuzushiji-MNIST and CIFAR-10. Table~\ref{tab:dataset} summarizes the benchmark datasets. Following \cite{lu19iclr}, we manually corrupted the 10-class datasets into binary classification datasets (see Appendix~\ref{sec:setup} for details). Two unlabeled training datasets $\Xtr$ and $\Xtr'$ of the same sample size are drawn according to Eq.~\eqref{eq:train-density}. And the risk is evaluated on them during training. Test data are just drawn from $p(x,y)$ for evaluations.

\begin{table*}[t]
    \caption{Means (standard deviations) of the classification accuracy (Acc) and the drop ($\Delta_A$) over five trials in percentage with deep models. The best and comparable methods based on the paired \textit{t}-test at the significance level 5\% are highlighted in boldface.}
    \label{tab:deep}
    \begin{center}\scriptsize
    \resizebox{170mm}{!}{
    \begin{tabular*}{0.855\textwidth}{cccccccccccc}
            \toprule
            \multirow{2}{*}{Dataset} & \multirow{2}{*}{$\theta$, $\theta'$} & \multicolumn{2}{c}{UU-Biased} & \multicolumn{2}{c}{UU-Unbiased} & \multicolumn{2}{c}{UU-ABS} & \multicolumn{2}{c}{UU-ReLU} & \multicolumn{2}{c}{UU-LReLU}\\
            \cmidrule(r){3-4} \cmidrule(r){5-6} \cmidrule(r){7-8} \cmidrule(r){9-10} \cmidrule(r){11-12}
            & & Acc & $\Delta_{A}$ & Acc & $\Delta_{A}$ & Acc & $\Delta_{A}$ & Acc & $\Delta_{A}$ & Acc & $\Delta_{A}$\\
            \midrule
            
            \multirow{6}{*}{MNIST} & \multirow{2}{*}{0.8, 0.2} & 80.56 & 14.99 & 78.01 & 18.07 & \bf95.19 & 0.86 & \bf95.15 & 1.11 & \bf95.21 & 1.06\\
            & & (0.62) & (0.74) & (0.45) & (0.55) & \bf(0.12) & (0.13) & \bf(0.43) & (0.36) & \bf(0.42) & (0.42)\\
            & \multirow{2}{*}{0.7, 0.3} & 70.55 & 20.80 & 64.74 & 29.78 & 91.69 & 2.77 & \bf93.01 & 1.88 & \bf93.29 & 1.60\\
            & & (0.66) & (0.75) & (0.78) & (0.84) & (1.13) & (1.08) & \bf(0.39) & (0.53) & \bf(0.81) & (0.76)\\
            & \multirow{2}{*}{0.6, 0.4} & 59.85 & 19.37 & 53.34 & 37.74 & 78.54 & 11.73 & \bf88.11 & 3.37 & \bf90.34 & 1.13\\
            & & (0.59) & (1.11) & (0.88) & (1.31) & (1.19) & (1.27) & \bf(1.48) & (1.38) & \bf(0.84) & (0.66)\\
            \midrule
            
            \multirow{6}{*}{\tabincell{c}{Fashion-\\MNIST}} & \multirow{2}{*}{0.8, 0.2} & 81.51 & 9.56 & 80.19 & 11.40 & \bf90.41 & 1.13 & 90.20 & 1.35 & \bf90.90 & 0.64\\
            & & (0.77) & (0.65) & (0.81) & (0.81) & \bf(0.56) & (0.43) & (0.53) & (0.37) & \bf(0.26) & (0.25)\\
            & \multirow{2}{*}{0.7, 0.3} & 72.07 & 16.23 & 71.93 & 18.48 & 87.84 & 2.43 & \bf88.14 & 2.22 & \bf89.39 & 0.97\\
            & & (0.94) & (0.63) & (1.42) & (1.29) & (0.80) & (0.70) & \bf(0.90) & (1.00) & \bf(0.18) & (0.15)\\
            & \multirow{2}{*}{0.6, 0.4} & 61.58 & 17.89 & 63.01 & 25.10 & 80.86 & 6.83 & 83.78 & 4.11 & \bf86.25 & 1.63\\
            & & (1.30) & (0.75) & (1.07) & (1.17) & (1.38) & (1.59) & (1.00) & (0.84) & \bf(0.32) & (0.24) \\
            \midrule
            
            \multirow{6}{*}{\tabincell{c}{Kuzushiji-\\MNIST}} & \multirow{2}{*}{0.8, 0.2} & 78.10 & 8.22 & 74.60 & 14.76 & \bf86.62 & 2.85 & \bf87.13 & 2.28 & \bf87.56 & 1.85\\
            & & (0.69) & (0.83) & (0.71) & (0.77) & \bf(1.11) & (1.19) & \bf(0.99) & (0.87) & \bf(0.62) & (0.45)\\
            & \multirow{2}{*}{0.7, 0.3} & 70.77 & 10.95 & 66.40 & 21.12 & 83.79 & 3.81 & \bf85.35 & 2.20 & \bf85.65 & 1.60\\
            & & (0.58) & (0.63) & (0.49) & (0.48) & (0.66) & (0.55) & \bf(0.60) & (0.41) & \bf(0.29) & (0.23)\\
            & \multirow{2}{*}{0.6, 0.4} & 61.70 & 11.44 & 60.12 & 23.59 & 77.82 & 5.79 & 80.52 & 3.32 & \bf82.22 & 1.61\\
            & & (0.76) & (1.41) & (0.90) & (1.12) & (1.12) & (1.19) & (1.35) & (1.10) & \bf(0.52) & (0.27)\\
            \midrule
            
            \multirow{6}{*}{\tabincell{c}{CIFAR-\\10}} & \multirow{2}{*}{0.8, 0.2} & 74.28 & 10.76 & 76.12 & 11.48 & \bf84.39 & 3.22 & \bf84.47 & 3.18 & \bf84.51 & 3.11\\
            & & (0.94) & (1.37) & (3.51) & (3.21) & \bf(1.34) & (1.04) & \bf(1.68) & (1.32) & \bf(1.33) & (0.95) \\
            & \multirow{2}{*}{0.7, 0.3} & 65.06 & 14.09 & 67.52 & 17.84 &\bf80.53 & 4.84 & \bf81.64 & 3.73 & \bf81.26 & 4.08\\
            & & (0.46) & (1.59) & (3.07) & (2.85) & \bf(1.52) & (0.72) & \bf(1.46) & (1.19) & \bf(2.51) & (2.44)\\
            & \multirow{2}{*}{0.6, 0.4} & 57.12 & 12.52 & 57.26 & 23.95 & 71.53 & 9.30 & 76.83 & 4.10 & \bf78.34 & 2.62\\
            & & (0.46) & (2.25) & (1.33) & (1.35) & (1.40) & (0.66) & (1.26) & (0.91) & \bf(1.00) & (0.69)\\
            
            \bottomrule
        \end{tabular*}}
    \end{center}
\end{table*}

\paragraph{Baselines}
In order to analyze the proposed methods, we compare them with two baselines:
\begin{itemize}
    \item \emph{UU-Biased} means supervised classification taking the U set with larger class prior as P data and the other U set with smaller class prior as N data, which is a straightforward method to handle UU classification problem. In our setup, two U sets are of the same sample size, thus UU-biased method reduces to the BER minimization method \citep{menon15icml};
    \item \emph{UU-Unbiased} means the state-of-the-art UU method proposed in \cite{lu19iclr}.
\end{itemize}
For our proposed methods, \emph{UU-ABS}, \emph{UU-ReLU}, \emph{UU-LReLU} are short for the unbiased UU method using ABSolute function, ReLU function and generalized Leaky ReLU function as consistent correction function in Eq.~\eqref{eq:nn-emp-risk-uu-hat} respectively.

\paragraph{Experimental setup}
We first demonstrate that the aforementioned overfitting problem cannot be solved by simply applying the regularization techniques in deep learning, such as dropout and weight decay. For space reasons, we defer the experimental results on general-purpose regularization and discussions to Appendix~\ref{sec:suppexp}. We then test our proposed methods under different training class prior settings: $(\theta, \theta')$ are chosen as $(0.9, 0.1)$, $(0.8, 0.2)$, and $(0.7, 0.3)$; and using different models which are summarized in Table~\ref{tab:dataset}: MLP refers to \emph{multi-layer perceptron}, ResNet refers to \emph{residual networks} \citep{he16cvpr} and their detailed architectures are in Appendix~\ref{sec:setup}.

We implemented all the methods by Keras\footnote{\url{https://keras.io}}, and conducted the experiments on a NVIDIA Tesla P100 GPU. As a common practice, Adam \citep{kingma15iclr} with logistic loss $\elllog(z)=\ln(1+\exp(-z))$ was used for optimization. We trained 200 epochs and besides the final classification accuracy~(Acc) we also report the classification accuracy drop~($\Delta_A$), which is the difference between the best accuracy of all training epochs and the accuracy at the end of training, to demonstrate overfitting. Note that for fair comparison, we use the same models and hyperparameters (see Appendix~\ref{sec:setup}) for the implementation of all methods.


\paragraph{Experimental results with simple models}
We firstly test on simple models and report our results in Table~\ref{tab:simple}. We can see that the UU-Unbiased method and three consistent risk correction methods outperform the UU-biased method. The advantage increases as the classification task becomes harder, that is, the class priors move closer\footnote{
Intuitively, as the class priors move closer, two U sets would be more similar and thus less informative, which is significantly harder than assuming that they are sufficiently far away.}. Moreover, the overfitting issue in UU-Unbiased method is not severe for linear models, but we can see the tendency that overfitting gets slightly worse when class priors are closer.

\paragraph{Experimental results with deep models}
We now test on deep neural networks which are more flexible than the aforementioned simple models. Our observations of the experimental results in Table~\ref{tab:deep} are as follows. First, compared to simple model experiments, the overfitting of the UU-biased and UU-Unbiased methods become catastrophic: the performance drops behind their linear counterparts. This may be explained by that flexible models have larger capacity to fit patterns (making negative partial risks $-b\hRui^-(g)$ and $-c\hRuii^+(g)$ in \eqref{eq:risk-uu-hat} as negative as possible) and thus the empirical risk tends to be negative. And we observe that the closer the class priors are, the more severe the overfitting is. Second, the proposed consistent risk correction methods significantly alleviate the overfitting even in the hardest learning scenario, and their classification accuracy improves compared to the simple model experiments. Among all methods, the UU-LReLU method achieves the best performance for all the datasets and class prior settings, and has the smallest performance drop when the class priors get closer, which implies that it is relatively robust against the closeness of class priors. 


\vspace{-0.5ex}%
\section{Conclusions}
\label{sec:concl}%
\vspace{-0.5ex}%
We focused on mitigating the overfitting problem of the state-of-the-art unbiased UU method. Based on our empirical observations, we conjecture the negative empirical training risk as a potential reason for the overfitting and proposed a correction method that wraps the false positive and false negative parts of the empirical risk in a family of consistent correction functions. Furthermore, we proved the consistency of the proposed risk estimators and their minimizers. Experiments demonstrated the superiority of our proposed methods, especially for using flexible neural network models.

\newpage
\subsubsection*{Acknowledgments}
We thank Aditya Krishna Menon, Takashi Ishida, Ikko Yamane and Wenkai Xu for helpful discussions. NL was supported by the MEXT scholarship No.\ 171536. MS was supported by JST CREST Grant Number JPMJCR18A2.

\bibliography{practical_uu}

\clearpage
\onecolumn
\begin{center}
\LARGE Supplementary Material for Mitigating Overfitting in
Supervised Classification from Two Unlabeled Datasets:\\
A Consistent Risk Correction Approach
\end{center}{}

\appendix
\section{Proofs}
\label{sec:proof}%

In this appendix, we prove all theorems.

\subsection{Proof of Lemma~\ref{thm:pr-diff}}
\label{subsec:Lemma2}%
Let
\begin{align*}
\ptr(\Xtr)=\ptr(x_1)\cdots \ptr(x_n),\quad
\ptr'(\Xtr')=\ptr'(x'_1)\cdots \ptr'(x'_{n'})
\end{align*}
be the probability density functions of $\Xtr$ and $\Xtr'$ (due to the i.i.d. sample assumption). Then, the measure of $\fD^-(g)$ is defined by
\begin{align*}
\pr(\fD^-(g)) = \int_{(\Xtr,\Xtr')\in\fD^-(g)}\ptr(\Xtr)\ptr'(\Xtr')\dif\Xtr\dif\Xtr',
\end{align*}
where $\pr$ denotes the probability, $\dif\Xtr=\dif x_1\cdots\dif x_n$ and $\dif\Xtr'=\dif x'_1\cdots\dif x'_{n'}$. Since $\hRuu(g)$ is unbiased and $\hRnnuu(g)-\hRuu(g)=0$ on $\fD^+(g)$, the bias of $\hRnnuu(g)$ can be formulated as:
\begin{align*}
\bE[\hRnnuu(g)]-R(g)
&= \bE[\hRnnuu(g)-\hRuu(g)]\\
&= \int_{(\Xtr,\Xtr')\in\fD^+(g)}\left(\hRnnuu(g)-\hRuu(g)\right)\,\ptr(\Xtr)\ptr'(\Xtr')\dif\Xtr\dif\Xtr'\\
&\quad +\int_{(\Xtr,\Xtr')\in\fD^-(g)}\left(\hRnnuu(g)-\hRuu(g)\right)\, \ptr(\Xtr)\ptr'(\Xtr')\dif\Xtr\dif\Xtr'\\
&= \int_{(\Xtr,\Xtr')\in\fD^-(g)}\left(\hRnnuu(g)-\hRuu(g)\right)\, \ptr(\Xtr)\ptr'(\Xtr')\dif\Xtr\dif\Xtr'\\
\end{align*}
Thus we have $\bE[\hRnnuu(g)]-R(g)>0$ if and only if $\int_{(\Xtr,\Xtr')\in\fD^-(g)}\ptr(\Xtr)\ptr'(\Xtr')\dif\Xtr\dif\Xtr'>0$ due to the fact that $\hRnnuu(g)-\hRuu(g)>0$ on $\fD^-(g)$. That is, the bias of $\hRnnuu(g)$ is positive if and only if the measure of $\fD^-(g)$ is non-zero.

Next we study the probability measure of $\fD^-(g)$ by \emph{the method of bounded differences}. Since $\Rp^+(g)\ge\alpha_{g}/\pip$ and $\Rn^-(g)\ge\beta_{g}/\pin$, then
\begin{align*}
\bE[A-C]=\pip\Rp^+(g)\ge\alpha_{g},\quad
\bE[D-B]=\pin\Rn^-(g)\ge\beta_{g}.
\end{align*}
We have assumed that $0\le\ell(z)\le C_\ell$, and thus the change of $a\hRui^+(g)$ and $b\hRui^-(g)$ will be no more than $aC_\ell/n$ and $bC_\ell/n$ if some $x_i\in\Xtr$ is replaced, or the change of $c\hRuii^+(g)$ and $d\hRuii^-(g)$ will be no more than $cC_\ell/n'$ and $dC_\ell/n'$ if some $x'_j\in\Xtr'$ is replaced. Subsequently, \emph{McDiarmid's inequality} \citep{mcdiarmid89MBD} implies
\begin{align*}
\pr\{\pip\Rp^+(g)-(A-C)\ge\alpha_{g}\}
&\le \exp\left(-\frac{2\alpha_{g}^2}{n(aC_\ell/n)^2+n'(cC_\ell/n')^2}\right)\\
&= \exp\left(-\frac{2\alpha_{g}^2/C_\ell^2}{a^2/n+c^2/n'}\right),
\end{align*}
and
\begin{align*}
\pr\{\pin\Rn^-(g)-(D-B)\ge\beta_{g}\}
&\le \exp\left(-\frac{2\beta_{g}^2}{n'(dC_\ell/n')^2+n(bC_\ell/n)^2}\right)\\
&= \exp\left(-\frac{2\beta_{g}^2/C_\ell^2}{b^2/n'+d^2/n}\right).
\end{align*}
Then the probability measure of $\fD^-(g)$ can be bounded by
\begin{align*}
\pr(\fD^-(g)) &\le \pr\{A-C\le0\}+\pr\{D-B<0\}\\
&\le \pr\{A-C\le\pip\Rp^+(g)-\alpha_{g}\}+\pr\{D-B\le\pin\Rn^-(g)-\beta_{g}\}\\
&= \pr\{\pip\Rp^+(g)-(A-C)\ge\alpha_{g}\}+\pr\{\pin\Rn^-(g)-(D-B)\ge\beta_{g}\}\\
&\le \exp\left(-\frac{2\alpha_{g}^2/C_\ell^2}{a^2/n+c^2/n'}\right)+\exp\left(-\frac{2\beta_{g}^2/C_\ell^2}{b^2/n'+d^2/n}\right),
\end{align*}
we complete the proof. \qed

\subsection{Proof of Theorem~\ref{thm:bias-consistency}}
Based on Lemma~\ref{thm:pr-diff}, we can show the exponential decay of the bias and also the consistency of the proposed non-negative risk estimator $\hRnnuu(g)$. It has been proved in Lemma~\ref{thm:pr-diff} that
\begin{align*}
\bE[\hRnnuu(g)]-R(g)=\int_{(\Xtr,\Xtr')\in\fD^-(g)}\left(\hRnnuu(g)-\hRuu(g)\right)\, \ptr(\Xtr)\ptr'(\Xtr')\dif\Xtr\dif\Xtr'.
\end{align*}
Therefore the exponential decay of the bias can be obtained via
\begin{align*}
\bE[\hRnnuu(g)]-R(g)
&\le \sup\nolimits_{(\Xtr,\Xtr')\in\fD^-(g)}\left(\hRnnuu(g)-\hRuu(g)\right)\cdot\int_{(\Xtr,\Xtr')\in\fD^-(g)}\ptr(\Xtr)\ptr'(\Xtr')\dif\Xtr\dif\Xtr'\\
&= \sup\nolimits_{(\Xtr,\Xtr')\in\fD^-(g)}(f_1(A-C)+f_2(D-B)-(A-C)-(D-B))\cdot\pr(\fD^-(g))\\
&\le \sup\nolimits_{(\Xtr,\Xtr')\in\fD^-(g)}(|f_1(A-C)|+|f_2(D-B)|+|A-C|+|D-B|)\cdot\pr(\fD^-(g))\\
&\le \sup\nolimits_{(\Xtr,\Xtr')\in\fD^-(g)}(L_f|A-C|+L_f|D-B|+|A-C|+|D-B|)\cdot\pr(\fD^-(g))\\
&= \sup\nolimits_{(\Xtr,\Xtr')\in\fD^-(g)}((L_f+1)|A-C|+(L_f+1)|D-B|)\cdot\pr(\fD^-(g))\\
&\le \sup\nolimits_{(\Xtr,\Xtr')\in\fD^-(g)}((L_f+1)(a+c)C_\ell+(L_f+1)(d+b)C_\ell)\cdot\pr(\fD^-(g))\\
&= (L_f+1)(a+b+c+d)C_\ell\Delta_{g},
\end{align*}
where we employed the Lipschitz condition, i.e., $|f_1(x)-f_1(y)|\le L_f|x-y|$ (also holds for $f_2$), and the assumption $f(0)=0$ in Definition~\ref{def:cc}. Then the deviation bound \eqref{eq:dev-bound} is due to
\begin{align*}
|\hRnnuu(g)-R(g)|
&\le |\hRnnuu(g)-\bE[\hRnnuu(g)]|+|\bE[\hRnnuu(g)]-R(g)|\\
&\le |\hRnnuu(g)-\bE[\hRnnuu(g)]|+(L_f+1)(a+b+c+d)C_\ell\Delta_{g}.
\end{align*}
Denote by $A'$, $B'$, $C'$ and $D'$ that differs from $A$, $B$ ,$C$ and $D$ on a single example. Then
\begin{align}
& |f_1(A-C)+f_2(D-B)-f_1(A'-C)-f_2(D-B')|\notag\\
&\quad \le |f_1(A-C)-f_1(A'-C)|+|f_2(D-B)-f_2(D-B')|\notag\\
&\quad \le L_f|A-C-A'+C|+L_f|D-B-D+B'|\notag\\
\label{eq:rcc1}
&\quad = L_f|A-A'|+L_f|B'-B|\notag\\
&\quad \le (a+b)L_fC_\ell/n.
\end{align}
Similarily, we can obtain
\begin{align}
\label{eq:rcc2}
|f_1(A-C)+f_2(D-B)-f_1(A-C')-f_2(D'-B)|
&\le (c+d)L_fC_\ell/n'.
\end{align}
Therefore the change of $\hRnnuu(g)$ will be no more than $(a+b)L_fC_\ell/n$ if some $x_i\in\Xtr$ is replaced, or it will be no more than $(c+d)L_fC_\ell/n'$ if some $x_j'\in\Xtr'$ is replaced, and McDiarmid's inequality gives us
\begin{align*}
\pr\{|\hRnnuu(g)-\bE[\hRnnuu(g)]|\ge\epsilon\}
\le 2\exp\left(-\frac{2\epsilon^2}{n((a+b)L_fC_\ell/n)^2+n'((c+d)L_fC_\ell/n')^2}\right).
\end{align*}
Setting the above right-hand side to be equal to $\delta$ and solving for $\epsilon$ yields immediately the following bound. For any $\delta>0$, the following inequality holds with probability at least $1-\delta$,
\begin{align*}
|\hRnnuu(g)-\bE[\hRnnuu(g)]|
&\le \sqrt{\frac{\ln(2/\delta)C_\ell^2L_f^2}{2}\left(\frac{(a+b)^2}{n}+\frac{(c+d)^2}{n'}\right)}\\
&\le C_\delta\left(\frac{(a+b)}{\sqrt{n}}+\frac{(c+d)}{\sqrt{n'}}\right)\\
&= C_\delta\cdot\chi_{n,n'},
\end{align*}
where $C_\delta=C_\ell L_f\sqrt{\ln(2/\delta)/2}$ and $\chi_{n,n'}=(a+b)/\sqrt{n}+(c+d)/\sqrt{n'}$.
Thus we obtain
\begin{align*}
|\hRnnuu(g)-R(g)| \le C_\delta\cdot\chi_{n,n'}+(L_f+1)(a+b+c+d)C_\ell\Delta_{g}.
\end{align*}
On the other hand, the deviation bound \eqref{eq:dev-bound-alter} is due to
\begin{align*}
|\hRnnuu(g)-R(g)| \le |\hRnnuu(g)-\hRuu(g)|+|\hRuu(g)-R(g)|,
\end{align*}
where $|\hRnnuu(g)-\hRuu(g)|>0$ with probability at most $\Delta_{g}$, and $|\hRuu(g)-R(g)|$ shares the same concentration inequality with $|\hRnnuu(g)-\bE[\hRnnuu(g)]|$. \qed

\subsection{Proof of Theorem~\ref{thm:est-err}}
First, we introduce the definitions of Rademacher complexity.
\begin{definition}[Rademacher complexity]
\label{def:rademacher}%
Let $\cG=\{g:\mathcal{Z}\to\bR\}$ be a class of measurable functions, $\cX=\{x_1,\ldots,x_n\}$ be a fixed sample of size $n$ i.i.d. drawn from a probability distribution $p$, and $\bm{\varepsilon}=(\varepsilon_1,\ldots,\varepsilon_n)^{T}$ be Rademacher variables, i.e., independent uniform random variables taking values in $\{-1,+1\}$. For any integer $n\ge1$, the Rademacher complexity of $\cG$ \citep{mohri12FML,ssbd14UML} is defined as
\begin{align*}
\fR_{n,p}(\cG) = \bE_\cX\bE_{\bm{\varepsilon}}\left[ \sup\nolimits_{g\in\cG}
\frac{1}{n}\sum\nolimits_{x_i\in\cX}\varepsilon_{i}g(x_i) \right].
\end{align*}
An alternative definition of the Rademacher complexity \citep{koltchinskii01tit,bartlett02jmlr} will be used in the proof is:
\begin{align*}
\fR'_{n,p}(\cG) = \bE_\cX\bE_{\bm{\varepsilon}}\left[ \sup\nolimits_{g\in\cG}
\left|\frac{1}{n}\sum\nolimits_{x_i\in\cX}\varepsilon_{i}g(x_i)\right| \right].
\end{align*}
\end{definition}

Then, we list all the lemmas that will be used to derive the estimation error bound in Theorem~\ref{thm:est-err}.
\begin{lemma}
\label{thm:lemma1}%
For arbitrary $\cG$, $\fR'_{n,p}(\cG)\ge\fR_{n,p}(\cG)$; if $\cG$ is closed under negation, $\fR'_{n,p}(\cG)=\fR_{n,p}(\cG)$.
\end{lemma}

\begin{lemma}[Theorem 4.12 in \cite{ledoux91PBS}]
\label{thm:lemma2}%
If $\psi:\bR\to\bR$ is a Lipschitz continuous function with a Lipschitz constant $L_\psi$ and satisfies $\psi(0)=0$, we have
\begin{align*}
\fR'_{n,p}(\psi\circ\cG) &\le 2L_\psi\fR'_{n,p}(\cG),
\end{align*}
where $\psi\circ\cG=\{\psi\circ g|g\in\cG\}$ and $\circ$ is a composition operator.
\end{lemma}

\begin{lemma}
\label{thm:lemma3}%
Under the assumptions of Theorem~\ref{thm:est-err}, for any $\delta>0$, with probability at least $1-\delta$,
  \begin{align}
  \sup\nolimits_{g\in\cG}|\hRnnuu(g)-R(g)|
  &\le 4(a+b)L_fL_\ell\fR_{n,\ptr}(\cG)+4(c+d)L_fL_\ell\fR_{n',\ptr'}(\cG)\notag\\
  \label{eq:uni-dev-bound}%
  &+(L_f+1)(a+b+c+d)C_\ell\Delta+C_\delta'\cdot\chi_{n,n'}.
  \end{align}
\begin{proof}
Firstly, we deal with the bias of $\hRnnuu(g)$. Noticing that the assumptions $\inf_{g\in\cG}\Rp^+(g)\ge\alpha/\pip>0$ and $\inf_{g\in\cG}\Rn^-(g)\ge\beta/\pin>0$ imply $\Delta=\sup_{g\in\cG}\Delta_{g}$. By \eqref{eq:bias-bound} we have:
\begin{align}
\sup\nolimits_{g\in\cG}|\hRnnuu(g)-R(g)|
&\le \sup\nolimits_{g\in\cG}|\hRnnuu(g)-\bE[\hRnnuu(g)]|
+\sup\nolimits_{g\in\cG}|\bE[\hRnnuu(g)]-R(g)| \notag\\
\label{eq:uni-dev-bias}%
&\le \sup\nolimits_{g\in\cG}|\hRnnuu(g)-\bE[\hRnnuu(g)]|
+(L_f+1)(a+b+c+d)C_\ell\Delta.
\end{align}
Secondly, we consider the double-sided uniform deviation $\sup\nolimits_{g\in\cG}|\hRnnuu(g)-\bE[\hRnnuu(g)]|$. Denote by $\cX_\mathrm{s}=\{(\Xtr,\Xtr')\}$, and $\cX_\mathrm{s}'$ that differs from $\cX_\mathrm{s}$ on a single example. Then we have
\begin{align*}
&\big|\sup\nolimits_{g\in\cG}\big|\hRnnuu(g;\cX_\mathrm{s})-\bE_{\cX_\mathrm{s}}[\hRnnuu(g;\cX_\mathrm{s})]\big|
-\sup\nolimits_{g\in\cG}\big|\hRnnuu(g;\cX_\mathrm{s}')-\bE_{\cX_\mathrm{s}'}[\hRnnuu(g;\cX_\mathrm{s}')]\big|\big|\\
&\quad \le \sup\nolimits_{g\in\cG}\big||\hRnnuu(g;\cX_\mathrm{s})-\bE_{\cX_\mathrm{s}}[\hRnnuu(g;\cX_\mathrm{s})]|
-|\hRnnuu(g;\cX_\mathrm{s}')-\bE_{\cX_\mathrm{s}'}[\hRnnuu(g;\cX_\mathrm{s}')]|\big|\\
&\quad \le \sup\nolimits_{g\in\cG}|\hRnnuu(g;\cX_\mathrm{s})-\hRnnuu(g;\cX_\mathrm{s}')|,
\end{align*}
where we applied the \emph{triangle inequality}. According to \eqref{eq:rcc1} and \eqref{eq:rcc2}, we see that the change of $\sup\nolimits_{g\in\cG}|\hRnnuu(g)-\bE[\hRnnuu(g)]|$ will be no more than $(a+b)L_fC_\ell/n$ if some $x_i\in\Xtr$ is replaced, or it will be no more than $(c+d)L_fC_\ell/n'$ if some $x_j'\in\Xtr'$ is replaced. Similar to the proof technique of Theorem~\ref{thm:bias-consistency}, by applying McDiarmid's inequality to the uniform deviation we have with probability at least $1-\delta$,
\begin{align}
&\sup\nolimits_{g\in\cG}|\hRnnuu(g)-\bE[\hRnnuu(g)]|-\bE[\sup\nolimits_{g\in\cG}|\hRnnuu(g)-\bE[\hRnnuu(g)]|]\notag\\
&\quad \le \sqrt{\frac{\ln(1/\delta)C_\ell^2L_f^2}{2}\left(\frac{(a+b)^2}{n}+\frac{(c+d)^2}{n'}\right)}\notag\\
\label{eq:uni-dev-martingale}
&\quad = C_\delta'\cdot\chi_{n,n'},
\end{align}
where $C_\delta'=C_\ell L_f\sqrt{\ln(1/\delta)/2}$.
Thirdly, we make \emph{symmetrization} \citep{vapnik98SLT}. Suppose that $(\Xtr^{gh},\Xtr^{'gh})$ is a \emph{ghost sample}, then
\begin{align*}
&\bE[\sup\nolimits_{g\in\cG}|\hRnnuu(g)-\bE[\hRnnuu(g)]|]\\
&\quad = \bE_{(\Xtr,\Xtr')}[\sup\nolimits_{g\in\cG}|\hRnnuu(g;\Xtr,\Xtr')-\bE_{(\Xtr^{gh},\Xtr^{'gh})}\hRnnuu(g;\Xtr^{gh},\Xtr^{'gh})|]\\
&\quad \le \bE_{(\Xtr,\Xtr')} [\sup\nolimits_{g\in\cG}\bE_{(\Xtr^{gh},\Xtr^{'gh})}|\hRnnuu(g;\Xtr,\Xtr')-\hRnnuu(g;\Xtr^{gh},\Xtr^{'gh})|]\\
&\quad \le \bE_{(\Xtr,\Xtr'),(\Xtr^{gh},\Xtr^{'gh})}[\sup\nolimits_{g\in\cG}|\hRnnuu(g;\Xtr,\Xtr')-\hRnnuu(g;\Xtr^{gh},\Xtr^{'gh})|],
\end{align*}
where we applied \emph{Jensen's inequality} twice since the absolute value and the supremum are convex. By decomposing the difference $|\hRnnuu(g;\Xtr,\Xtr')-\hRnnuu(g;\Xtr^{gh},\Xtr^{'gh})|$, we can know that
\begin{align*}
&|\hRnnuu(g;\Xtr,\Xtr')-\hRnnuu(g;\Xtr^{gh},\Xtr^{'gh})|\\
&\quad = \Big|f_1\left(a\hRui^+(g;\Xtr)-c\hRuii^+(g;\Xtr')\right)- f_1\left(a\hRui^+(g;\Xtr^{gh})-c\hRuii^+(g;\Xtr^{'gh})\right)\\
&\quad + f_2\left(-b\hRui^-(g;\Xtr)+d\hRuii^-(g;\Xtr')\right)- f_2\left(-b\hRui^-(g;\Xtr^{gh})+d\hRuii^-(g;\Xtr^{'gh})\right)\Big|\\
&\quad \le\Big|f_1\left(a\hRui^+(g;\Xtr)-c\hRuii^+(g;\Xtr')\right)- f_1\left(a\hRui^+(g;\Xtr^{gh})-c\hRuii^+(g;\Xtr^{'gh})\right)\Big|\\
&\quad + \Big|f_2\left(-b\hRui^-(g;\Xtr)+d\hRuii^-(g;\Xtr')\right)- f_2\left(-b\hRui^-(g;\Xtr^{gh})+d\hRuii^-(g;\Xtr^{'gh})\right)\Big|\\
&\quad \le\Big|L_f\left(a\hRui^+(g;\Xtr)-c\hRuii^+(g;\Xtr')-a\hRui^+(g;\Xtr^{gh})+c\hRuii^+(g;\Xtr^{'gh})\right)\Big|\\
&\quad + \Big|L_f\left(-b\hRui^-(g;\Xtr)+d\hRuii^-(g;\Xtr')+b\hRui^-(g;\Xtr^{gh})-d\hRuii^-(g;\Xtr^{'gh})\right)\Big|\\
&\quad \le\Big|aL_f\left(\hRui^+(g;\Xtr)-\hRui^+(g;\Xtr^{gh})\right)\Big| + \Big|cL_f\left(\hRuii^+(g;\Xtr')-\hRuii^+(g;\Xtr^{'gh})\right)\Big|\\
&\quad + \Big|bL_f\left(\hRui^-(g;\Xtr)-\hRui^-(g;\Xtr^{gh})\right)\Big| + \Big|dL_f\left(\hRuii^-(g;\Xtr')-\hRuii^-(g;\Xtr^{'gh})\right)\Big|,
\end{align*}
where we employed the Lipschitz condition. This decomposition results in
\begin{align*}
\bE[\sup\nolimits_{g\in\cG}|\hRnnuu(g)-\bE[\hRnnuu(g)]|]
&\le aL_f\bE_{\Xtr,\Xtr^{gh}}\Big[\sup\nolimits_{g\in\cG}\Big|\left(\hRui^+(g;\Xtr)-\hRui^+(g;\Xtr^{gh})\right)\Big|\Big]\\
&\quad + cL_f\bE_{\Xtr',\Xtr^{'gh}}\Big[\sup\nolimits_{g\in\cG}\Big|\left(\hRuii^+(g;\Xtr')-\hRuii^+(g;\Xtr^{'gh})\right)\Big|\Big]\\
&\quad + bL_f\bE_{\Xtr,\Xtr^{gh}}\Big[\sup\nolimits_{g\in\cG}\Big|\left(\hRui^-(g;\Xtr)-\hRui^-(g;\Xtr^{gh})\right)\Big|\Big]\\
&\quad +dL_f\bE_{\Xtr',\Xtr^{'gh}}\Big[\sup\nolimits_{g\in\cG}\Big|\left(\hRuii^-(g;\Xtr')-\hRuii^-(g;\Xtr^{'gh})\right)\Big|\Big].
\end{align*}
Fourthly, we relax those expectations to Rademacher complexities. The original $\ell$ may miss the origin, i.e., $\ell(0,y)\neq0$, with which we need to cope. Let
\begin{align*}
\bar{\ell}(t,y)=\ell(t,y)-\ell(0,y)
\end{align*}
be a \emph{shifted loss} so that $\bar{\ell}(0,y)=0$. Hence,
\begin{align*}
\hRui^+(g;\Xtr)-\hRui^+(g;\Xtr^{gh})
&\textstyle = (1/n)\sum_{x_i\in\Xtr}\ell(g(x_i),+1)
-(1/n)\sum_{x^{gh}_i\in\Xtr^{gh}}\ell(g(x^{gh}_i),+1)\\
&\textstyle = (1/n)\sum_{i=1}^n(\ell(g(x_i),+1)-\ell(g(x^{gh}_i),+1))\\
&\textstyle = (1/n)\sum_{i=1}^n(\bar{\ell}(g(x_i),+1)-\bar{\ell}(g(x^{gh}_i),+1)).
\end{align*}
This is already a standard form where we can attach Rademacher variables to every $\bar{\ell}(g(x_i),+1)-\bar{\ell}(g(x^{gh}_i),+1)$, so we have
\begin{align*}
&\bE_{\Xtr,\Xtr^{gh}}[\sup\nolimits_{g\in\cG}|\hRui^+(g;\Xtr)-\hRui^+(g;\Xtr^{gh})|]\\
&\quad = \bE_{\Xtr,\Xtr^{gh}}\Big[\sup\nolimits_{g\in\cG}\Big|(1/n)\sum_{i=1}^n(\bar{\ell}(g(x_i),+1)-\bar{\ell}(g(x^{gh}_i),+1))\Big|\Big]\\
&\quad = \bE_{\bm{\varepsilon},\Xtr,\Xtr^{gh}}\Big[\sup\nolimits_{g\in\cG}\Big|(1/n)\sum_{i=1}^n\varepsilon_{i}(\bar{\ell}(g(x_i),+1)-\bar{\ell}(g(x^{gh}_i),+1))\Big|\Big]\\
&\quad \le \bE_{\bm{\varepsilon},\Xtr}\Big[\sup\nolimits_{g\in\cG}\Big|(1/n)\sum_{i=1}^n\varepsilon_{i}(\bar{\ell}(g(x_i),+1)\Big|\Big]\\
&\quad + \bE_{\bm{\varepsilon},\Xtr^{gh}}\Big[\sup\nolimits_{g\in\cG}\Big|(1/n)\sum_{i=1}^n\varepsilon_{i}(\bar{\ell}(g(x^{gh}_i),+1)\Big|\Big]\\
&\quad = 2\bE_{\bm{\varepsilon},\Xtr}\Big[\sup\nolimits_{g\in\cG}\Big|(1/n)\sum_{i=1}^n\varepsilon_{i}(\bar{\ell}(g(x_i),+1)\Big|\Big]\\
&\quad = 2\fR'_{n,\ptr}(\bar{\ell}(\cdot,+1)\circ\cG)
\end{align*}
The other three expectations can be handled analogously. As a result,
\begin{align*}
\bE[\sup\nolimits_{g\in\cG}|\hRnnuu(g)-\bE[\hRnnuu(g)]|]
&\le 2aL_f\fR'_{n,\ptr}(\bar{\ell}(\cdot,+1)\circ\cG)+2cL_f\fR'_{n',\ptr'}(\bar{\ell}(\cdot,+1)\circ\cG) \\
&\quad +2bL_f\fR'_{n,\ptr}(\bar{\ell}(\cdot,-1)\circ\cG)+2dL_f\fR'_{n',\ptr'}(\bar{\ell}(\cdot,-1)\circ\cG).
\end{align*}
Finally, we transform the Rademacher complexities of composite function classes to the original function class. It is obvious that $\bar{\ell}$ shares the same Lipschitz constant $L_\ell$ with $\ell$, and consequently

\begin{align*}
\fR'_{n,\ptr}(\bar{\ell}(\cdot,+1)\circ\cG)&\le 2L_\ell\fR'_{n,\ptr}(\cG)= 2L_\ell\fR_{n,\ptr}(\cG) \\
\fR'_{n',\ptr'}(\bar{\ell}(\cdot,+1)\circ\cG)&\le 2L_\ell\fR'_{n',\ptr'}(\cG)= 2L_\ell\fR_{n',\ptr'}(\cG) \\
\fR'_{n,\ptr}(\bar{\ell}(\cdot,-1)\circ\cG)&\le 2L_\ell\fR'_{n,\ptr}(\cG)= 2L_\ell\fR_{n,\ptr}(\cG) \\
\fR'_{n',\ptr'}(\bar{\ell}(\cdot,-1)\circ\cG)&\le 2L_\ell\fR'_{n',\ptr'}(\cG)= 2L_\ell\fR_{n',\ptr'}(\cG) \\
\end{align*}
where we used the assumption that $\cG$ is closed under negation, Lemma~\ref{thm:lemma1} and Lemma~\ref{thm:lemma2}. So we have
\begin{align}
\label{eq:uni-dev-rademacher-normal}%
\bE[\sup\nolimits_{g\in\cG}|\hRnnuu(g)-\bE[\hRnnuu(g)]|]
&\le 4(a+b)L_fL_\ell\fR_{n,\ptr}(\cG)+4(c+d)L_fL_\ell\fR_{n',\ptr'}(\cG).
\end{align}
Combining \eqref{eq:uni-dev-bias}, \eqref{eq:uni-dev-martingale} and \eqref{eq:uni-dev-rademacher-normal} finishes the proof of the uniform deviation bound \eqref{eq:uni-dev-bound}.
\end{proof}
\end{lemma}

We are now ready to prove our estimation error bound based on the uniform deviation bound in Lemma~\ref{thm:lemma3}.
\begin{align*}
R(\hgcc)-R(g^*)
&= \left(\hRnnuu(\hgcc)-\hRnnuu(g^*)\right)
+\left(R(\hgcc)-\hRnnuu(\hgcc)\right)
+\left(\hRnnuu(g^*)-R(g^*)\right)\\
&\quad \le 0 +2\sup\nolimits_{g\in\cG}|\hRnnuu(g)-R(g)|\\
&\quad \le 8(a+b)L_fL_\ell\fR_{n,\ptr}(\cG)+8(c+d)L_fL_\ell\fR_{n',\ptr'}(\cG)\\
&\quad +2(L_f+1)(a+b+c+d)C_\ell\Delta+2C_\delta'\cdot\chi_{n,n'},
\end{align*}
where $\hRnnuu(\hgcc)\le\hRnnuu(g^*)$ by the definition of $g^*$ and $\hgcc$. \qed

\subsection{Proof of Corollary~\ref{thm:deep-bound}}
We further get bounds on the Rademacher complexity of deep neural networks by the following Theorem.

\begin{theorem}[Theorem 1 in \cite{golowich2017size}]
\label{thm:colla}%
Assume the Frobenius norm of the weight matrices $W_j$ are at most $M_{F}(j)$, and the activation function $\sigma$ satisfying the assumption that it is $1$-Lipschitz, positive-homogeneous which is applied element-wise (such as the ReLU). Let $x$ is upper bounded by $C_x$. Then,
  \begin{align}
  \fR_{n,p}(\cG)
  \le \frac{1}{n} \prod_{j=1}^{m} M_{F}(j) \cdot(\sqrt{2m\log 2 }+1) \sqrt{\sum_{i=1}^{n}\left\|\mathbf{x}_{i}\right\|^{2}}
  \le \frac{C_x(\sqrt{2m\log 2}+1) \prod_{j=1}^{m} M_{F}(j)}{\sqrt{n}}.
  \end{align}
\end{theorem}

Based on Theorem~\ref{thm:colla}, we proved
\begin{align*}
R(\hgcc)-R(g^*)
&\le 8(a+b)L_fL_\ell\fR_{n,\ptr}(\cG)+8(c+d)L_fL_\ell\fR_{n',\ptr'}(\cG)\\
&\quad +2(L_f+1)(a+b+c+d)C_\ell\Delta+2C_\delta'\cdot\chi_{n,n'}\\
&\quad \le 8(a+b)L_fL_\ell\frac{C_x(\sqrt{2m\log 2}+1) \prod_{j=1}^{m} M_{F}(j)}{\sqrt{n}}\\
&\quad +8(c+d)L_fL_\ell\frac{C_x(\sqrt{2m\log 2}+1) \prod_{j=1}^{m} M_{F}(j)}{\sqrt{n'}}\\
&\quad +2(L_f+1)(a+b+c+d)C_\ell\Delta+2C_\delta'\cdot\chi_{n,n'}\\
&\quad = \left(8L_fL_\ell C_x(\sqrt{2m\log 2}+1) \prod_{j=1}^{m} M_{F}(j)+2C_\delta'\right)\cdot\chi_{n,n'}\\
&\quad +2(L_f+1)(a+b+c+d)C_\ell\Delta.
\end{align*}
\section{Supplementary information on Figure~\ref{fig:illustration}}
\label{sec:supp_figure1}%

In Sec.~\ref{sec:unbiased}, we illustrated the overfitting issue of state-of-the-art unbiased UU method using different datasets, different models, different optimizers and different loss functions. The details of these demonstration results are presented here.

In the upper row, the dataset used was MNIST and we artifically corrupt it into a binary classification dataset: even digits form the P class and odd digits form the N class. The models used were a linear-in-input model~(Linear) $g(x)=\boldsymbol{\omega}^T x+b$ where $\boldsymbol{\omega}\in\bR^{784}$ and $b\in\bR$, and a 5-layer \emph{multi-layer perceptron}~(MLP): $d$-300-300-300-300-1. And the optimizer was SGD with momentum (momentum=0.9) with logistic loss $\elllog(z)=\ln(1+\exp(-z))$ or sigmoid loss $\ellsig(z)=1/(1+\exp(z))$. For linear model experiments, the batch size was fixed to be $1000$ and the initial learning rate was $5e-2$. For MLP model experiments, the batch size was fixed to be $3000$ and the initial learning rate was $1e-3$.

In the bottom row, the dataset used was CIFAR-10 and we artifically corrupt it into a binary classification dataset: the P class is composed of `bird', `cat', `deer', `dog', `frog', and `horse', and the N class is composed of `airplane', `automobile', `ship' and `truck'. The models used were \emph{all convolutional net}~(AllConvNet) \citep{springenberg15iclr} as follows:
\begin{itemize}[leftmargin=10em]
    \item[0th (input) layer:] (32*32*3)-
    \item[1st to 3rd layers:] [C(3*3, 96)]*2-C(3*3, 96, 2)-
    \item[4th to 6th layers:] [C(3*3, 192)]*2-C(3*3, 192, 2)-
    \item[7th to 9th layers:] C(3*3, 192)-C(1*1, 192)-C(1*1, 10)-
    \item[10th to 12th layers:] 1000-1000-1
\end{itemize}
where C(3*3, 96) means 96 channels of 3*3 convolutions followed by ReLU, [ $\cdot$ ]*2 means 2 such layers, C(3*3, 96, 2) means a similar layer but with stride 2, etc; and a 32-layer \emph{residual networks}~(ResNet32) \citep{he16cvpr} as follows:
\begin{itemize}[leftmargin=10em]
    \item[0th (input) layer:] (32*32*3)-
    \item[1st to 11th layers:] C(3*3, 16)-[C(3*3, 16), C(3*3, 16)]*5-
    \item[12th to 21st layers:] [C(3*3, 32), C(3*3, 32)]*5-
    \item[22nd to 31st layers:] [C(3*3, 64), C(3*3, 64)]*5-
    \item[32nd layer:] Global Average Pooling-1
\end{itemize}
where [ $\cdot$, $\cdot$ ] means a building block \citep{he16cvpr}. Batch normalization \citep{ioffe15icml} was applied before hidden layers. An $\ell_2$-regularization was added, where the regularization parameter was fixed to 5e-3. The models were trained by Adam \citep{kingma15iclr} with the default momentum parameters ($\beta_1=0.9$ and $\beta_2=0.999$) and the loss function was $\ellsig(z)$ or $\elllog(z)$. For AllConvNet experiments, the batch size and the learning rate were fixed to be $500$ and $1e-5$ respectively. For ResNet32 experiments, the batch size and the learning rate were fixed to be $3000$ and $3e-5$ respectively.

The two training distributions were created following \eqref{eq:train-density} with class priors $\theta=0.6$ and $\theta'=0.4$. Subsequently, the two sets of U training data were sampled from those distributions with sample sizes $n=30000$ and $n'=30000$.

The results demonstrate the concurrence of empirical training risk going negative (blue dashed line) and the test accuracy overfitting (green dashed line) regardless of datasets, models, optimizers and loss functions. 

\section{Supplementary information on the experiments}
\label{sec:setup}%

\paragraph{MNIST \citep{lecun98mnist}} This is a grayscale image dataset of handwritten digits from 0 to 9 where the size of the images is 28*28. It contains 60,000 training images and 10,000 test images. See \url{http://yann.lecun.com/exdb/mnist/} for details. Since it has 10 classes originally, we used the even digits as the P class and the odd digits as the N class, respectively.

The simple model used for training MNIST was a linear-in-input model $g(x)=\boldsymbol{\omega}^T x+b$ where $\boldsymbol{\omega}\in\bR^{784}$ and $b\in\bR$ with $\ell_2$-regularization (the regularization parameter was fixed to be $1e-4$). The batch size and learning rate were set to be 3000 and $1e-3$ respectively. The deep model uased was a 5-layer FC with ReLU as the activation function: $d$-300-300-300-300-1 with $\ell_2$-regularization (the regularization parameter was fixed to be $5e-3$). The batch size and learning rate were set to be 3000 and $5e-5$ respectively. For both models, batch normalization \citep{ioffe15icml} with the default $momentum=0.99$ and $\epsilon=1e-3$ was applied before hidden layers, and the model was trained by Adam with the default momentum parameters $\beta_1=0.9$ and $\beta_2=0.999$. For all the experiments, the generalized leaky ReLU hyperparameter $\lambda$ was selected from $-0.01$ to $1$.

\paragraph{Fashion-MNIST \citep{xiao17arxiv}} This is also a grayscale fashion image dataset similarly to MNIST, but here each data is associated with a label from 10 fashion item classes. See \url{https://github.com/zalandoresearch/fashion-mnist} for details. It was converted into a binary classification dataset as follows:
\begin{itemize}
    \item the P class is formed by `T-shirt', `Trouser', `Shirt', and `Sneaker';
    \item the N class is formed by `Pullover', `Dress', `Coat', `Sandal', `Bag', and `Ankle boot'.
\end{itemize}
The models and optimizers were the same as MNIST, where the learning rate for the simple and deep models were set to be $5e-3$ and $3e-5$ and the other hyperparameters remain the same.

\paragraph{Kuzushiji-MNIST \citep{clanuwat2018deep}} This is another variant of MNIST dataset consisting of 60,000 training images and 10,000 test images of cursive Japanese (Kuzushiji) characters. See \url{https://github.com/rois-codh/kmnist} for details. For Kuzushi-MNIST dataset,
\begin{itemize}
    \item `ki', `re', `wo' made up the P class;
    \item `o', `su', `tsu', `na', `ha', `ma', `ya' made up the N class.
\end{itemize}
The models and optimizers were the same as MNIST, where the learning rate for the deep models was set to be $3e-5$ and the other hyperparameters remain the same.

\paragraph{CIFAR-10 \citep{krizhevsky09cifar}} This dataset consists of 60,000 $32*32$ color images in 10 classes, and there are 5,000 training images and 1,000 test images per class. See \url{https://www.cs.toronto.edu/~kriz/cifar.html} for details. For CIFAR-10 dataset,
\begin{itemize}
    \item the P class is composed of `bird', `cat', `deer', `dog', `frog' and `horse';
    \item the N class is composed of `airplane', `automobile', `ship' and `truck'.
\end{itemize}

The simple model used for training CIFAR-10 was also a linear-in-input model $g(x)=\boldsymbol{\omega}^T x+b$ where $\boldsymbol{\omega}\in\bR^{3072}$ and $b\in\bR$ with $\ell_2$-regularization (the regularization parameter was fixed to be $5e-3$). The batch size and learning rate were set to be 3000 and $5e-3$ respectively. The deep model was again ResNet-32 \citep{he16cvpr} that can be find in Appendix~\ref{sec:supp_figure1}. The batch size and learning rate were set to be $3000$ and $5e-3$ respectively. For both models, batch normalization \citep{ioffe15icml} with the default $momentum=0.99$ and $\epsilon=1e-3$ was applied before hidden layers, and the model was trained by Adam with the default momentum parameters $\beta_1=0.9$ and $\beta_2=0.999$.

\section{Supplementary experiments on general-purpose regularization}
\label{sec:suppexp}%
Regularization is the most common technique that lower the complexity of a neural network model during training, and thus prevent the overfitting \citep{goodfellow16DL}. In this section, we demonstrate that the general-purpose regularization methods fail to mitigate the overfitting in UU classification scenario.

We tested the unbiased UU method using two most popular regularization techniques, i.e., dropout and weight decay. The dataset and model used were again MNIST (the class priors $\theta$ and $\theta'$ were set to be $0.6$ and $0.4$) and the 5-layer MLP $d$-300-300-300-300-1 with $\ell_2$-regularization, where dropout layers were added between the existing layers. The optimizer was SGD with momentum ($momentum=0.9$) and logistic loss. Batch size was fixed to be $3000$ and the initial learning rate was $1e-3$. For dropout experiments, we fixed the weight decay parameter to be $1e-4$ and change the dropout parameter from $0$ to $0.8$. For weight decay experiments, we fix the dropout parameter to be $0.2$ and change the weight decay parameter from $0.0005$ to $5$.

\begin{figure*}[t]
    \centering
    \includegraphics[width=0.45\textwidth]{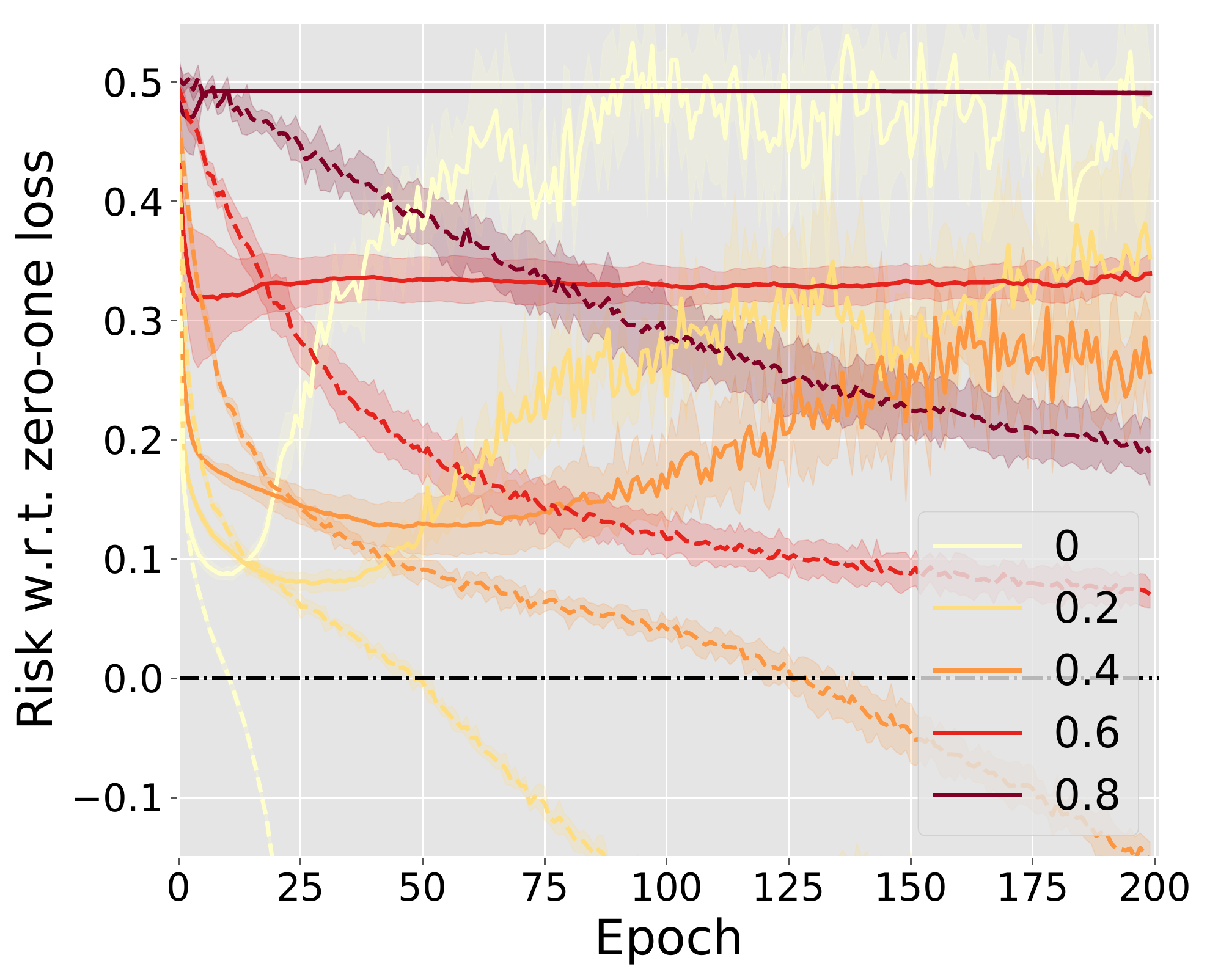}
    \includegraphics[width=0.45\textwidth]{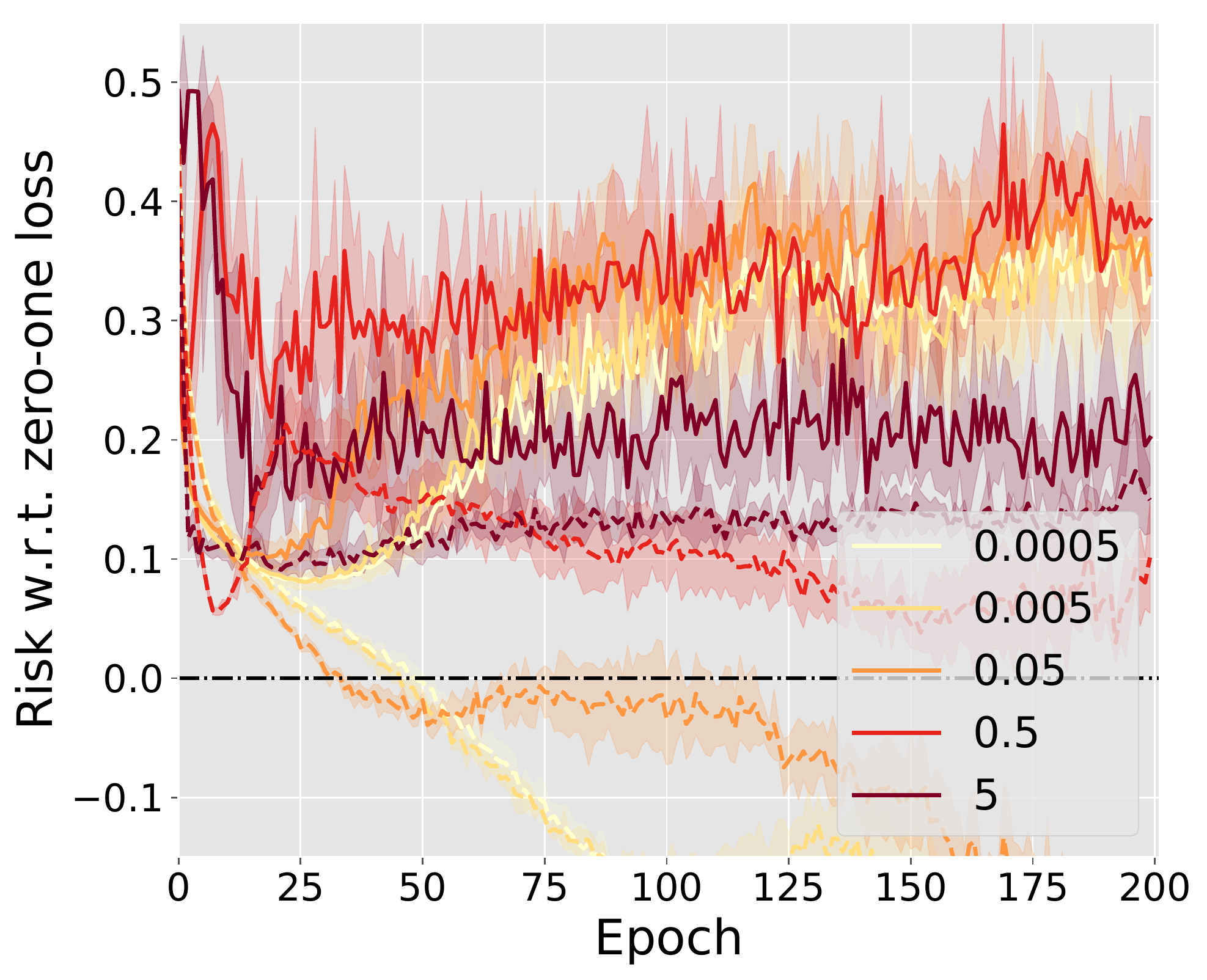}
    \caption{Supplementary experimental results on general regularization. Left: dropout. Right: weight decay. Solid curves are $\hRuu(g)$ on test data and dashed curves are $\hRuu(g)$ on training data.}
    \label{fig:suppreg}
\end{figure*}

Empirical results in Figure~\ref{fig:suppreg} show that the unbiased UU method with slightly strong regularizations outperforms the one with weaker regularizations, but still suffers from overfitting. It is because adding strong regularization may prohibit the high representation power of deep models, which in turn may cause underfitting. 

\paragraph{Discussion} Instead of the general-purpose regularization, our proposed method explicitly utilizes the additional knowledge that the empirical risk goes negative. By that, we can more "effectively" constrain the model without too much sacrificing the representation power of deep models. Note that our correction can also be regarded as a regularization in its general sense for fighting against overfitting, but differently from weight decay or dropout, it is exclusively designed for UU classification and hence it is no surprising that our regularization fits UU classification better than other regularizations as discussed in Sec.~\ref{sec:pratical_uu} theoretically and demonstrated in Sec.~\ref{sec:exp} empirically.

\end{document}